\documentclass[final,12pt]{clear2026} 

\usepackage[most]{tcolorbox}
\usepackage{bm}

\usepackage{aligned-overset}
\usepackage{xspace}
\usepackage{xcolor}
\usepackage{mathrsfs}
\usepackage{booktabs}
\usepackage{pifont}

\newcommand{\xmark}{\ding{55}}
\usepackage{thmtools}
\usepackage{amssymb}
\usepackage{makecell}
\usepackage{acronym}
\usepackage{float}

\usepackage{tikz} 

\newcommand{\R}{\mathbb{R}}                

\newcommand{\Normal}[2]{\mathcal{N} \left(#1, #2 \right)}

\newcommand{\E}{\mathbb{E}}

\newcommand{\ind}{\mathrel{\perp\!\!\!\perp}}
\newcommand{\nind}{\mathrel{\not\!\perp\!\!\!\perp}}

\DeclareMathOperator{\Unif}{Unif}

\newtheorem{scm}{SCM}
\newtheorem{claim}{Claim}
\newcommand{\Z}{\bm{Z}}
\newcommand{\z}{\bm{z}}
\newcommand{\A}{\bm{A}}
\renewcommand{\a}{\bm{a}}
\newcommand{\Y}{\bm{Y}}
\newcommand{\y}{\bm{y}}

\newcommand{\Anoise}{\bm{U_{\a}}}
\newcommand{\anoise}{\bm{u_{\a}}}
\newcommand{\Ahatnoise}{\bm{\tilde{U}_{\a}}}
\newcommand{\ahatnoise}{\bm{\tilde{u}_{\a}}}
\newcommand{\Ynoise}{\bm{U_{\y}}}
\newcommand{\ynoise}{\bm{u_{\y}}}
\newcommand{\Yhatnoise}{\bm{\tilde{U}_{\y}}}
\newcommand{\yhatnoise}{\bm{\tilde{u}_{\y}}}
\newcommand{\Znoise}{\bm{U_{\z}}}

\newcommand{\Zhatnoise}{\bm{\tilde{U}_{\z}}}
\newcommand{\zhatnoise}{\bm{\tilde{u}_{\z}}}
\newcommand{\Noisehat}{\bm{\tilde{U}}}
\newcommand{\noisehat}{\bm{\tilde{u}}}

\newcommand{\Yhat}{\tilde{\Y}}

\newcommand{\Ahat}{\tilde{\A}}

\newcommand{\Zhat}{\tilde{\Z}}

\newcommand{\ynoiseuniv}{u_y}

\newcommand{\yhatnoiseuniv}{\tilde{u}_y}

\newcommand{\Yhatnoisecond}{(\Yhatnoise \mid \Z = \z, \A = \a)}
\newcommand{\Ynoisecond}{(\Ynoise \mid \Z = \z, \A = \a)}

\newcommand{\Ycond}{(\Y \mid \Z = \z, \A = \a)}

\newcommand{\Zdomain}{\mathcal{\bm{Z}}}
\newcommand{\Adomain}{\mathcal{\bm{A}}}
\newcommand{\Ydomain}{\mathcal{\bm{Y}}}

\newcommand{\Ynoisedomain}{\mathcal{\bm{U}}_y}

\newcommand{\Anoisedomain}{\mathcal{\bm{U}}_a}

\title[Flow IV For Nonseparable Counterfactual Inference]{Flow IV: Counterfactual Inference In Nonseparable Outcome Models Using Instrumental Variables}
\usepackage{times}

\clearauthor{%
 \Name{Marc Braun} \Email{marc.braun@liu.se}\\
 \addr Department of Computer and Information Science, Linköping University
 \AND
 \Name{Jose M. Peña} \Email{jose.m.pena@liu.se}\\
 \addr Department of Computer and Information Science, Linköping University%
 \AND
  \Name{Adel Daoud} \Email{adel.daoud@liu.se}\\
 \addr Institute for Analytical Sociology, Linköping University\\
 Chalmers University of Technology%
}

\begin{document}
\newacro{DGP}{data-generating process}
\newacro{IV}{instrumental variable}
\newacro{SCM}{structural causal model}
\newacro{pdf}{probability density function}
\newacro{MLP}{multiplayer perceptron}

\maketitle

\begin{abstract}%
  To reach human level intelligence, learning algorithms need to incorporate causal reasoning. But identifying causality, and particularly counterfactual reasoning, remains elusive. In this paper, we make progress on counterfactual inference in nonseparable outcome models by utilizing \acp{IV}. \acp{IV} are a classic tool for mitigating bias from unobserved confounders when estimating causal effects. While \ac{IV} methods for effect estimation have been extended to nonseparable outcome models under different assumptions, existing \ac{IV} approaches to counterfactual prediction typically assume one-dimensional outcomes and additive noise. In this paper, we show that under standard \ac{IV} assumptions, along with the assumption that the outcome function is invertible and has a triangular structure, then the treatment–outcome relationship becomes identifiable from observed data. We furthermore propose a method to learn the outcome function utilizing normalizing flows. This outcome function estimator can then be used to perform counterfactual inference. We refer to the method as \textit{Flow IV}.%
\end{abstract}
\acresetall


\section{Introduction}

Estimating causal effects is a central goal in many scientific fields, including the social and medical sciences. Unlike traditional machine learning methods, which rely on associational relationships, causal inference aims to answer questions about the effect of an intervention, such as how physical exercise affects cholesterol levels. Randomized controlled trials are the gold standard for such questions, but ethical, logistical, or financial constraints often necessitate reliance on observational data.

Causal inference from observational data requires assumptions about the \ac{DGP}, most notably the absence of hidden confounders—unobserved variables that influence both treatment and outcome \citep{Hernan2025,lin2023}. Denoting unobserved factors affecting the treatment and outcome by $\Anoise$ and $\Ynoise$, respectively, this assumption implies independence between these variables \citep{Pearl1995}.

While average treatment effects are informative, many applications require more granular conclusions. Counterfactual inference goes beyond average or even conditional average causal effects and targets individual-level outcomes by reasoning about latent characteristics. For example, knowing the cholesterol level and amount of exercise, one may ask what an individual’s cholesterol level would have been had they exercised a different amount. Counterfactual reasoning allows personalized decision-making in domains such as medicine, education, and policy \citep{Kino2021}. A formal definition of counterfactual inference is given in Section~\ref{sec:problem-setup}.

When hidden confounders are present, \acp{IV} provide a principled approach to causal identification. An instrument $\Z$ affects the outcome only through its influence on the treatment. While classical \ac{IV} methods address linear settings \citep{Wald1940, Angrist1996}, more recent work extends \ac{IV} approaches to non-linear settings \citep{Chesher2003, Imbens2009, Guo2016, Puli2020}. However, existing \ac{IV}-based approaches to counterfactual inference typically assume separable outcome functions and one-dimensional outcomes \citep{Hartford2017, Singh2019}.

In practice, outcome functions are often nonseparable. For instance, the effect of age on cholesterol may differ depending on the level of exercise. In this work, we introduce a new identifiability result for counterfactuals under the assumptions that (i) a strong instrument is available and (ii) the outcome function follows a known triangular structure and is strictly monotonic in $\Ynoise$. Building on this result, we propose \textit{Flow IV}, a method based on normalizing flows or flow matching that flexibly models complex, high-dimensional outcome functions beyond additive noise assumptions.

The remainder of the paper is organized as follows. Section~\ref{sec:background} introduces the problem setup, \ac{IV} assumptions, and related work. Section~\ref{sec:identifiability} presents our identifiability result, and Section~\ref{sec:method} describes the proposed Flow IV method. Experimental results, including a real-world application to aid allocation in Africa \citep{Dreher2021}, are presented in Section~\ref{sec:experiments}. Section~\ref{sec:discussion} concludes with a discussion and future directions.
\section{Background and Related Work}
\label{sec:background}

This section defines the general structure of the type of \acp{DGP} this paper aims to perform counterfactual inference for and reviews related work.

\subsection{Problem Setup}
\label{sec:problem-setup}

Let $\Z$, $\A$, and $\Y$ be continuous random variables of dimension $k$, $m$, and $n$ respectively. Let $\Ynoise$ be an unobserved continuous random variable of dimension $n$ and let $\Anoise$ be an unobserved random variable of dimension $w$. With $p_X$ we denote the \ac{pdf} of random variable $X$. We call $\A$ the treatment and $\Y$ the outcome. $\Z$ is called instrument or \ac{IV} if it satisfies the following three conditions where $\Y(\a,\z)$ is the potential outcome of $\Y$ under intervention $\A=\a,\ \Z=\z$.

\begin{enumerate}
\item \textit{Relevance condition} $\Z \nind \A$ states that the instrument is associated with the treatment,
\item \textit{exclusion restriction} $\Y(\z, \a) = \Y(\z', \a) \ \text{ for all } \z,\ \z'$ requires that the instrument has no direct effect on the outcome, and
\item \textit{marginal exchangeability} $\Y(\a, \z) \ind \Z \ \text{ for all } \z,\ \a$, requires that the instrument and outcome do not share common causes.
\end{enumerate}

Let the \ac{DGP} for $(\Z, \A, \Y)$ be defined by the following \ac{SCM} \citep{Pearl2009}.

\begin{scm}
\label{scm:setup}
\begin{align*}
    \Z &= f_{\z}(\Znoise) \\
    \A &= f_{\a}(\Z, \Anoise) \\
    \Y &= f_{\y}(\A, \Ynoise)
\end{align*}    
\end{scm}

where $\Znoise$ is independent of $\Anoise$ and $\Ynoise$. The joint distribution of $\Anoise$ and $\Ynoise$ represents possible latent confounding between $\A$ and $\Y$. The corresponding causal graph of this \ac{SCM} can be seen in Figure \ref{fig:iv-dag}.

\begin{figure}[ht]
\centering
\begin{tikzpicture}[
  node distance=1.5cm and 1.5cm,
  every node/.style={draw, circle, minimum size=0.8cm},
  every path/.style={->,>=stealth,thick},
  scale=2
]    
    \node[dashed] (eZ) at (0, 0.8) {$\Znoise$};
    \node[dashed] (eA) at (1, 0.8) {$\Anoise$};
    \node[dashed] (eY) at (2, 0.8) {$\Ynoise$};
    \node (Z) at (0, 0) {$\Z$};
    \node (A) at (1, 0) {$\A$};
    \node (Y) at (2, 0) {$\Y$};
    
    \draw[] (Z) -- (A);
    \draw[] (A) -- (Y);
    \draw[] (eZ) -- (Z);
    \draw[] (eA) -- (A);
    \draw[] (eY) -- (Y);
    \draw[-] (eA) -- (eY);

\end{tikzpicture}
\caption{Causal graph illustrating the structure of the type of \acp{DGP} considered in this paper.}
\label{fig:iv-dag}
\end{figure}

Note that, under this \ac{DGP}, $\Z$ is an \ac{IV} and the outcome function $f_{\y}$ can be any separable or nonseparable function. Formally, $f_{\y}$ is called separable if it can be written as $f_{\y}(\A, \Ynoise) = \Tilde{f}_{\y}(\A) + \Ynoise$ where $\Tilde{f}_{\y}$ is a possibly nonlinear function.

The goal of this paper is to perform counterfactual inference of the form "What would the value of $\Y_i$ for individual $i$ have been, had the value of $\A_i$ been at the unobserved level $\a'$ when we actually observe $\Y_i = \y,\ \A_i = \a$?" We express this counterfactual outcome formally as $\Y_i(\a') \mid \Y_i = \y,\ \A_i = \a$. Answering these types of counterfactual queries requires three steps \citep{Pearl2009}.
\begin{enumerate}
\item \textit{Abduction}, where we infer the latent value ${\Ynoise}_i$ from the observed $(\A_i, \Y_i)$,
\item \textit{action}, where we replace the value of $\A_i=\a$ with $\a'$, and
\item \textit{prediction}, where we estimate the counterfactual outcome as $\Y_i(\a') = f_{\y}(\a', {\Ynoise}_i)$.
\end{enumerate}

Note that while \textit{probabilistic} counterfactual inference is possible under some assumptions, the counterfactual outcome is \textit{uniquely} identified if and only if the following condition for the outcome function $f_{\y}$ holds for all $\a, \a', \ynoise$, and $\ynoise'$

\begin{align}
\label{cond:unique-cf}
    f_{\y}(\a, {\ynoise}) = f_{\y}(\a, {\ynoise}') \iff f_{\y}(\a', {\ynoise}) = f_{\y}(\a', {\ynoise}')
\end{align}

The condition says that if the abduction step results in a set $\{{\ynoise}, {\ynoise}'\}$ of possible values for $\Ynoise$ that produce the factual $\y$ given $\a$, then the counterfactual value of $\y$ under ${\ynoise}$ has to be the same as the counterfactual value of $\y$ under ${\ynoise}'$ for all alternative treatments $\a'$. Otherwise, the counterfactual is not uniquely identified. Note that when the condition is true, then we can rewrite \ac{SCM} \ref{scm:setup} such that $f_{\y}$ is invertible while the model stays counterfactually equivalent (proof in Appendix \ref{app:technical-proofs}).
\subsection{Related Work}
\label{sec:related-work}

Utilizing \acp{IV} to estimate causal effects has a long tradition \citep{Wald1940, Angrist1996}. The classical \ac{IV} estimator assumes a linear relationship between $\Z$, $\A$, and $\Y$ such that $\Y = \bm{\beta} \A + \Ynoise$. In the simplest setup with a single instrument, $\beta$ can then be estimated using the so called Wald estimator $\hat{\beta} = \mathrm{Cov}(Y, Z) / \mathrm{Cov}(A, Z)$ \citep{Wald1940}. Following \citet{Hartford2017}, we refer to it as the Usual \ac{IV} approach. When the linearity assumption is violated, the Usual \ac{IV} estimator can be used to estimate the complier average treatment effect \citep{Liu2025}. There is a vast body of literature that utilizes \acp{IV} to identify such causal effects for specific subpopulations (local average treatment effects) \citep{Imbens1994, Robins1994, Imbens2010}.

Other literature aims at finding a \textit{control function} $C(\Z, \A)$ such that conditioning on the control variable $\bm{V} \coloneq C(\Z, \A)$ renders the treatment $\A$ independent of $\Ynoise$ \citep{Chesher2003, Imbens2009, Guo2016, Hahn2017, Puli2020}. The control variable then blocks all backdoor paths relative to $(\A, \Y)$ which allows identification of causal effects. While Figure \ref{fig:iv-dag} shows that in the setup of this paper, a control variable exists in the form of $\Anoise$, the method we propose does not directly aim at finding a control function, but yields a control function as a by-product. \citet{Puli2020} propose what they call the \textit{general control function} method (GCFN). They learn a control function using an autoencoder, where the encoder learns a distribution $F_{\bm{V} \mid \Z, \A}$ and the decoder learns to reconstruct $\A$ from $\Z$ and $\bm{V}$. However, guaranteeing that $\bm{V}$ is a valid control function with the GCFN method requires knowledge about the structural form of the treatment function like additivity or multiplicity and that this structural form is reflected by the decoder. After training the control function, a neural network can be used to regress the outcome on the treatment and control function. This estimator of $\E \left[ \Y \mid \A = \a, \bm{V} = v \right]$ can then be used to estimate the average effect as $\E \left[ \Y \mid do(\A = \a) \right] = \E_{v \sim F_{\bm{V}}} \left[ \E \left[ \Y \mid do(\A = \a), \bm{V}=v \right] \right] = \E_{v \sim F_{\bm{V}}} \left[ \E \left[ \Y \mid \A = \a, \bm{V}=v \right] \right]$ as $\bm{V}$ blocks all backdoor paths. The distribution of $\bm{V}$ is obtained from the decoder and the observed data distribution. How control functions can be used for counterfactual inference with nonseparable outcome functions is not the scope of above mentioned papers.

There exists literature that explicitly targets counterfactual inference with \acp{IV} that has relaxed the linearity assumption. However, \citet{Hartford2017} and \citet{Singh2019} assume separable outcome functions and \citet{NasrEsfahany2023} assume that instrument and treatment are discrete. Note that separability of the outcome function is a sufficient condition for the (necessary and sufficient) condition for counterfactual inference in Equation~\ref{cond:unique-cf}.
\citet{Hartford2017} propose the \textit{Deep IV} framework, where the problem is divided into two prediction tasks. In the first stage, the conditional distribution of the treatment under the instrument is learned whereas the second stage network models $g_{\y}(\A) + \E[\Ynoise] $. To counteract the spurious correlations between $\A$ and $\Y$, their loss function when training the second network involves integrating over the treatment distribution conditioned on the instrument. Note that under the separability assumption, we can also use other methods such as control function approaches to perform counterfactual inference as the abduction and prediction becomes trivial.

Table~\ref{tab:literature_comparison} summarizes the additional assumptions (beyond the existence of an \ac{IV}) made by existing \ac{IV} approaches, and includes our Flow IV method, whose assumptions are formally introduced in Section~\ref{sec:identifiability}. The table also indicates whether an approach supports interventional inference, i.e., predicting outcome distributions under interventions for a (sub-) population, and counterfactual inference, i.e., predicting individual-level potential outcomes.

\begin{table}[ht]
\centering
\begin{tabular}{lccc}
\toprule
\textbf{Approach} & \textbf{Assumptions} & \textbf{Interv.} & \textbf{Counterf.} \\
\midrule
Usual \ac{IV} & Linear outcome function & \checkmark & \checkmark \\
Deep IV & Separable outcome function & \checkmark & \checkmark \\
GCFN & Structural treatment process assumptions & \checkmark & \xmark \\
\textbf{Flow IV (ours)} & Triangular and strictly monotonic outcome function & \checkmark & \checkmark \\
\bottomrule
\end{tabular}
\caption{Comparison of assumptions and capabilities across \ac{IV} approaches. In accordance with Pearl's causal hierarchy, the last two columns refer to the capability of performing interventional (Interv.) and counterfactual (Counterf.) inference. All approaches additionally require the existence of an \ac{IV}.}
\label{tab:literature_comparison}
\end{table}

\section{Identifiability of the Outcome Function}
\label{sec:identifiability}

In this section, we introduce a new identifiability result for counterfactual outcomes that relaxes the assumptions made in existing literature.
First, we formally state the two assumptions required for the identifiability result to hold.

\begin{restatable}[Triangular Monotonicity]{assumption}{ytriangularmonotonicity}
\label{ass:triangular-monotonic}
    For $f_{\y}$ in \ac{SCM} \ref{scm:setup} we assume that $\ynoise \mapsto f_{\y}(\a, \ynoise)$ is triangular, i.e.
    \begin{equation*}
    f_{\y}(\a, \ynoise) =
         \begin{pmatrix}
             f_{\y}^{(1)}(\a, u_{\y}^{(1)}) \\
             f_{\y}^{(2)}(\a, u_{\y}^{(2)}, u_{\y}^{(1)}) \\
             \vdots \\
             f_{\y}^{(n)}(\a, u_{\y}^{(n)}, u_{\y}^{(n-1)}, \dots, u_{\y}^{(1)})
         \end{pmatrix}
    \end{equation*}
    and that each $f_{\y}^{(i)}$ is strictly monotonic with respect to $u_{\y}^{(i)}$.
\end{restatable}

Note that the separability assumption is a special case of the triangular monotonicity assumption and we therefore strictly relax the assumptions of existing approaches. In the setting where $\Y$ is one-dimensional, triangular monotonicity is equivalent to invertibility. Then, Assumption~\ref{ass:triangular-monotonic} is a necessary and sufficient condition for unique counterfactual inference and is therefore not specific to our approach (recall Section~\ref{sec:problem-setup}). For one-dimensional outcomes, previous research has argued why monotonicity and thus invertibility in the confounder is not unreasonable to hold in reality \citep{Ogburn2012}. For higher-dimensional outputs, Assumption~\ref{ass:triangular-monotonic} is sufficient for unique counterfactual inference.

The triangularity assumption is inspired by triangular systems, which are common in econometrics, and is assumed to hold in real-world applications where the effect unfolds over time. In such cases, the outcome  $y^{(t)}$ at time $t$ is a function of the treatment and the hidden causes $\ynoise$ up to time $t$, i.e., $y^{(t)} = f_{\y}^{(t)}(\a, u_{\y}^{(t)}, u_{\y}^{(t-1)}, \dots, u_{\y}^{(1)})$ (see Figure~\ref{fig:temporal-dag} in Appendix~\ref{app:justification-of-a1} for causal graph).

\begin{restatable}[Strong IV]{assumption}{strongiv}
\label{ass:strong-instrument}
    We assume that $p_{\A \mid \Z}(\A=\a \mid \Z = \z) > 0$ for every $\a$ and $\z$ such that $p_{\A}(\A=\a) > 0$ and $p_{\Z}(\Z = \z) > 0$.
\end{restatable}

Assumption~\ref{ass:strong-instrument} is similarly used in other \ac{IV} literature and typically referred to as \textit{strong IV} assumption. It states that any value of the treatment can be observed given any value of the instrument with non-zero probability, which can be assessed by domain experts. This leads to the novel identifiability result in the following theorem.

\begin{theorem}[Identifiability]
    \label{thrm:identifiabiliy}
    Let Assumption~\ref{ass:triangular-monotonic} and \ref{ass:strong-instrument} be true. Let $g_{\y}: \R^m \times \R^n \rightarrow \R^n$ be a function where $\yhatnoise \mapsto g_{\y}(\a, \yhatnoise)$ is triangular monotonic following the same triangular structure as $f_{\y}$ and let $\Yhat \coloneq g_{\y}(\A, \Yhatnoise)$ be a random variable. Let $\Yhatnoise$ be a random variable that has the same independencies as $\Ynoise$ in the causal graph in Figure \ref{fig:iv-dag}.
    
    Then for every $g_{\y}$ such that $(\Yhat, \A, \Z) \overset{d}{=} (\Y, \A, \Z)$, we have that

    \begin{equation*}
        g_{\y}(\a, \yhatnoise) = f_{\y}(\a, \psi(\yhatnoise))
    \end{equation*}

    where $\psi$ is an invertible function.
\end{theorem}

The proof can be found in Appendix \ref{app:technical-proofs}. Theorem \ref{thrm:identifiabiliy} states that any such function $g_{\y}$ that produces the correct observed distribution differs from the true outcome function $f_{\y}$ only by an invertible transformation of the latent term $\yhatnoise$. This immediately implies that using the function $g_{\y}$ in the abduction and prediction steps results in the same counterfactual value as using the function $f_{\y}$ (proof in Appendix \ref{app:technical-proofs}). We can therefore use $g_{\y}$ to predict counterfactuals.

\section{Implementation of Flow IV}
\label{sec:method}

In this section we introduce \textit{Flow IV}, an approach that makes use of Theorem \ref{thrm:identifiabiliy} to identify counterfactuals under relaxed assumptions compared to existing literature. The method uses normalizing flows or flow matching to model an invertible mapping between a latent space $(\Zhatnoise, \Ahatnoise, \Yhatnoise)$ and the observed space $(\Z, \A, \Y)$. The training procedure as pseudocode can be found in Appendix~\ref{app:algorithms}.

According to Theorem \ref{thrm:identifiabiliy}, the latent variable $\Yhatnoise$ has to respect the same independencies as $\Ynoise$ and by definition of $g_{\y}$, we know that $\Yhat$ must follow the same independencies as $\Y$ in Figure \ref{fig:iv-dag}. We therefore specify the flow model such that it respect the same independencies specified in the causal graph in Figure \ref{fig:iv-dag}. Normalizing flows that respect the independencies specified in a causal graph have been used before and are referred to as causal graphical normalizing flows (cGNFs) \citep{Balgi2022, Javaloy2023, Balgi2024}. They learn a conditional normalizing flow $g_i$ from standard normal noise $\mathbf{U}_i$ to each observed variable $V_i$ conditioned on the parents of $V_i$ in the corresponding causal graph and thereby resemble autoregressive-flows \citep{Kingma2016}. The conditional flow transformations in our setup are the following.
\begin{align*}
    \Zhat &= g_{\z}(\Zhatnoise; \theta_{\z}) \\
    \Ahat &= g_{\a}(\Zhat, \Ahatnoise; \theta_{\a}) \\
    \Yhat &= g_{\y}(\Ahat, \Yhatnoise; \theta_{\y})
\end{align*}

Note how closely the conditional flow transformations resemble \ac{SCM} \ref{scm:setup}. In contrast to existing cGNF approaches that assume no hidden confounding, in our setup $\Yhatnoise$ and $\Ahatnoise$ need not be independent. We therefore introduce another normalizing flow $h: \R^m \times \R^n \rightarrow \R^m \times \R^n$ from independent standard normal variables $\mathbf{\varepsilon}_{\a}$ and $\mathbf{\varepsilon}_{\y}$ to model the joint distribution of $(\Ahatnoise, \Yhatnoise)$. We call the parameters of this transformation $\theta_{\noisehat}$. $\Zhatnoise$ is independent of the other noise terms. Let the distribution of $\Zhatnoise$ be the standard normal distribution.

\subsection{Flow IV with Normalizing Flows}
\label{section:method-normalizing-flows}

By using autoregressive normalizing flows, the structure of $g_{\y}$ can be chosen such that it has the same monotonic triangular structure as $f_{\y}$ as it is required by Theorem \ref{thrm:identifiabiliy}. The parameters $\theta = (\theta_{\z}, \theta_{\a}, \theta_{\y}, \theta_{\noisehat})$ of the transformations can be optimized via maximum likelihood estimation given observations $\mathcal{D} = \{(\z_i, \a_i, \y_i) \}_{i=1}^{\ell}$.
\begin{align*}
    \hat{\theta} &= \arg\max_{\theta} \log \mathcal{L}(\theta \mid \mathcal{D}) \\
    \log \mathcal{L}(\theta \mid \mathcal{D})
    &= \sum_{i=1}^\ell \log p_{\text{obs}}(\z_i, \a_i, \y_i \mid \theta) \\
    &= \sum_{i=1}^\ell \log p_{\Noisehat} \left( g^{-1}(\z_i, \a_i, \y_i; \theta) \mid \theta_{\noisehat} \right)  + \log | \det \mathbb{J}_{g^{-1}(\z_i, \a_i, \y_i; \theta)} |
\end{align*}

Here, $p_{\text{obs}}$ and $p_{\Noisehat}$ are the \acp{pdf} of the target variables $(\Zhat, \Ahat, \Yhat)$ and the source variable $\Noisehat = (\Zhatnoise, \Ahatnoise, \Yhatnoise)$ respectively, $g^{-1}(\z_i, \a_i, \y_i) = (g_{\z}^{-1}(\z_i), g_{\a}^{-1}(\z_i, \a_i), g_{\y}^{-1}(\a_i, \y_i))$ is the inverse transformation, and $\mathbb{J}_{g^{-1}(\z_i, \a_i, \y_i)}$ denotes the Jacobian matrix of $g^{-1}$ evaluated at $\z_i, \a_i, \y_i$.

As $(\Ahatnoise, \Yhatnoise)$ is modelled with another normalizing flow, the density $p_{\Noisehat}$ in the likelihood equation can be derived similarly to the density $p_{\text{obs}}$ above. With $\varphi$ being the \ac{pdf} of the standard normal distribution we have
\begin{align*}
    \log p_{\Noisehat}(\zhatnoise, \ahatnoise, \yhatnoise \mid \theta_{\noisehat}) &= \log \varphi(\zhatnoise) + \log p_{\Ahatnoise, \Yhatnoise} \left( \ahatnoise, \yhatnoise \mid \theta_{\noisehat} \right) \\
    &= \log \varphi(\zhatnoise) + \log \varphi (\mathbf{\varepsilon}_{\a}) + \log \varphi (\mathbf{\varepsilon}_{\z}) + \log | \mathbb{J}_{h^{-1} (\ahatnoise, \yhatnoise; \theta_{\noisehat})} | \\
    \text{with } (\mathbf{\varepsilon}_{\a}, \mathbf{\varepsilon}_y) &= h^{-1}(\ahatnoise, \yhatnoise; \theta_{\noisehat})
\end{align*}

Normalizing flows converge to the true data distribution for $\ell \rightarrow \infty$ if the transformations are sufficiently flexible \citep{Papamakarios2021}. According to our identifiability result in Theorem \ref{thrm:identifiabiliy} we can then use $g_{\y}$ to obtain consistent estimates of counterfactuals.

\subsection{Flow IV with Flow Matching}
\label{sec:method-flow-matching}

While normalizing flows can in theory be used to model distributions of any dimensionality, flow matching has been shown to be very effective for modelling high-dimensional distributions. The method builds on continuous normalizing flows, where the time dependent flow transformation $g_{\y}$ is expressed indirectly through a vector field and an ordinary differential equation (ODE) \citep{Lipman2023}. The exact ODE is defined later in this section.

It is easy to show that choosing a neural network architecture for the vector field that has the same triangular structure as the outcome function implies that the solution of the ODE (i.e. the function $g_{\y}$) will have the same triangular structure. Encoding such triangular structure in architectures like \acp{MLP} is straightforward, but how to do this for more complex architecture remains to be investigated by future research. We show empirically in Section \ref{sec:experiments} that Flow IV outperforms associational models even when the triangular structure is not enforced.

We propose a two step process for training a flow matching model for Flow IV.
First, optimize $\theta_{\z}, \theta_{\a}$ of the normalizing flows $g_{\Z}$ and $g_{\A}$ defined in the previous subsection via the maximum-likelihood objective and the observations $\{(\z_i, \a_i)\}_{i=1}^{\ell}$. For sufficiently flexible normalizing flow transformations, the generated variables $(\Zhat, \Ahat)$ then converge to the true distribution of $(\Z, \A)$ for $\ell \rightarrow \infty$.

In a second step, define the distribution of $(\Ahatnoise, \Yhatnoise)$ via a normalizing flow transformation as in the previous subsection. However, now the marginal distribution of $\Ahatnoise$ is already fixed in the first step to train the normalizing flow $g_{\A}$. We therefore use a conditional flow $h$ from $\mathbf{\varepsilon}_{\Y}$ to $\Yhatnoise$ conditioned on $\Ahatnoise$. We can do this without loss of generality since the marginal distribution of $\Ahatnoise$ does not restrict the expressivity of our model.

Then, optimize the parameters $\theta_{\noisehat}$ of $h$ jointly with the parameters $\theta_{\y}$ of the vector field $v_t$ via the conditional flow matching objective \citep{Lipman2023} by sampling a source sample $\Yhatnoise$ from the conditional normalizing flow $h$ conditioned on $\Ahatnoise = g_{\A}^{-1}(\z_i, \a_i)$ and matching it with $\y_i$ from the training set $\{(\z_i, \a_i, \y_i)\}_{i=1}^{\ell}$.

We thereby perform flow matching between the latent space $\Yhatnoise \mid \Ahatnoise$ at time $t=0$ and an observed space $\Yhat \mid \Ahat, \Zhat$ at time $t=1$. Note that in the setup of this paper, the flow generating the outcome variable is conditioned on the treatment variable $\Ahat$ to respect the given independencies. The vector field $v: [0, 1] \times \R^m \times \R^n \rightarrow \R^n$ then constructs $g_{\y}$ via the following ODE.
\begin{align*}
    \frac{\partial}{\partial t} g_{\y}(\a, \ynoise, t) &= v(t, \a, g_{\y}(\a, \ynoise, t)) \\
    g_{\y}(\a, \ynoise, 0) &= \ynoise
\end{align*}

To perform counterfactual inference, the backward time ODE from $t=1$ to $t=0$ can be numerically solved in the abduction step and the forward time ODE from $t=0$ to $t=1$ can be solved numerically in the prediction step. The flow matching model that generates $\Yhat \mid \Ahat, \Zhat$ converges to the true distribution of $\Y \mid \A, \Z$ for $\ell \rightarrow \infty$. Consequently, the transformation $g_{\y}$ implied by the vector field $v$ can be used to predict counterfactuals according to Theorem \ref{thrm:identifiabiliy}.
\section{Experiments}
\label{sec:experiments}

In this section, we evaluate our Flow IV method in different experimental settings with synthetic data by comparing it to the Deep IV and GCFN\footnote{The GCFN method does not aim at performing counterfactual inference. The simplest approach to enable it to perform counterfactual inference is to assume separable outcome functions which is what we do in our experiments in order to compare it to Flow IV which does not require separability.} approaches. Furthermore, we illustrate how Flow IV can be used to produce counterfactual predictions for high-dimensional outputs like images. Lastly, we reanalyse a study performed by \citet{Dreher2021} that investigates the impact of Chinese foreign aid on a region's economic development. Some additional experiments and details about the experimental setup can be found in the Appendix \ref{app:experiments}.

\subsection{Synthetic Data Experiments}

We consider three different synthetic \acp{DGP} where all variables are 1D.
In the first \ac{DGP}, the outcome function is separable and therefore triangular monotonic. We call it \ac{DGP} 1 and it satisfies the assumptions of Flow IV, Deep IV and GCFN.
In the second \ac{DGP}, $Y$ is generated from a nonseparable but invertible (in 1D equivalent to triangular monotonic) outcome function. Therefore, only the assumptions of Flow IV are satisfied.
In a third setup, we consider a noninvertible (and therefore also nonseparable) outcome function. Therefore, the assumptions of none of the approaches are fulfilled.

\begin{align*}
    \text{DGP 1:} \qquad  &Y = 0.6 \cdot A + U_y &U_y \propto \alpha \cdot U_a^2 + \frac{1}{8}\eta \quad\text{ with } U_a, \eta \overset{iid}{\sim} \mathcal{N}(0,1) \\
    \text{DGP 2:} \qquad &Y = \left(\sin{(A+1.5)} + 1 \right) \cdot U_y &U_y \propto \alpha \cdot U_a^2 + \frac{1}{8}\eta \quad\text{ with } U_a, \eta \overset{iid}{\sim} \mathcal{N}(0,1) \\
    \text{DGP 3:} \qquad &Y = 0.6 \cdot (A + U_y)^2 & (U_y, U_a) \sim \Normal{\mathbf{0}}{\begin{pmatrix}
        1 & \rho \\
        \rho & 1
    \end{pmatrix}} \\
    &&\quad\text{ with } \rho = - \exp(-\alpha) + 1
\end{align*}

All terms $U_y$ are scaled to have mean zero and standard deviation one. The variable $\alpha \in \R^+$ controls the strength of confounding between $A$ and $Y$. The full \acp{DGP} can be found in Appendix~\ref{app:experiments}.

As an evaluation metric, we use the expected squared prediction error for counterfactual outcomes or cf-MSE for short. We define it as follows.

\begin{align*}
    \text{cf-MSE} \coloneq \E_{(A, Y) \sim p_{A, Y}} \left[ \E_{A' \sim p_A} \left[ (Y(A') - \hat{Y}(A'))^2 \mid A, Y \right] \right]
\end{align*}

We draw 20{,}000 samples from these distributions and use Monte Carlo estimation to estimate the expected value. Note that this is only possible for synthetic data, because only then can we calculate the true counterfactual outcome $Y(a') \mid A=a, Y=y$. A low counterfactual MSE indicates good predictive quality for counterfactuals.

\begin{figure}[ht]
    \centering
    \includegraphics[width=0.99\linewidth]{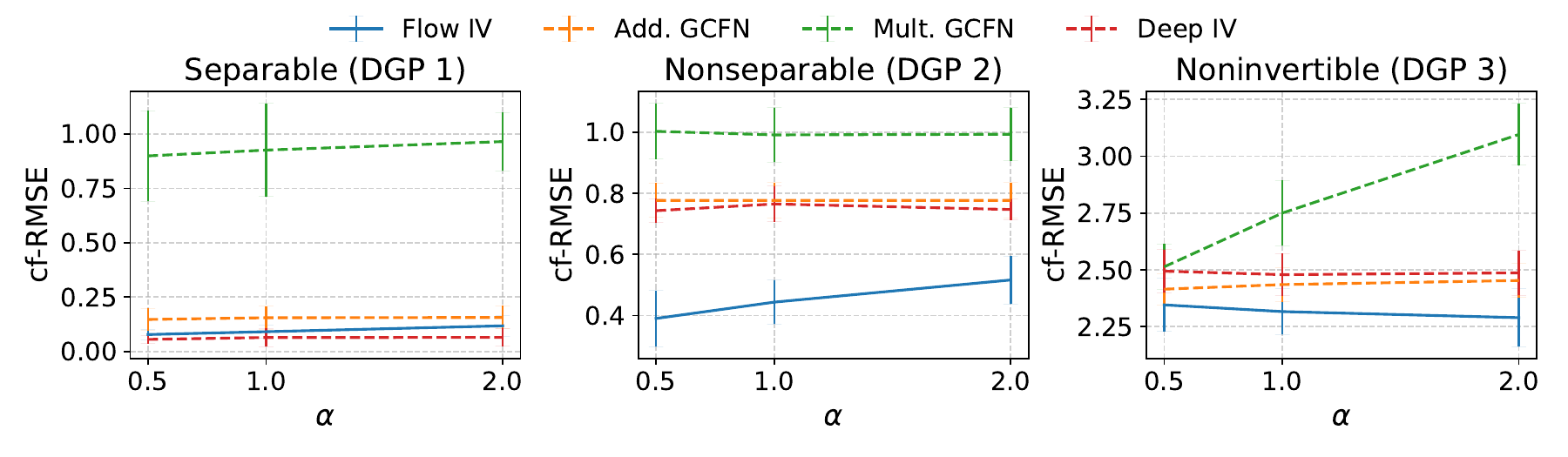}
    \caption{Comparison between Flow IV, Deep IV and GCFN in three different setups. Add. GCFN denotes the GCFN method with an additive decoder and Mult. GCFN with a multiplicative decoder. $\alpha$ controls the strength of confounding.}
    \label{fig:model-comparison}
\end{figure}

Figure \ref{fig:model-comparison} shows the estimated cf-RMSEs (defined as the square root of the cf-MSE) for the three different \acp{DGP}. 

We can see that Flow IV is on par or outperforms Deep IV and GCFN in all setups and for almost all levels of confounding that were tested. The difference is the largest for \ac{DGP} 2, because there, the triangular monotonicity assumption of Flow IV is satisfied whereas the separability assumption of Deep IV and GCFN is violated. According to the condition in Equation~\ref{cond:unique-cf} in Section~\ref{sec:problem-setup}, unique identification of counterfactuals is impossible for \ac{DGP} 3, implying that there exists no model that could achieve a cf-MSE of zero in this setup. Nonetheless, Flow IV outperforms existing methods in \ac{DGP} 3 which could be due to the higher flexibility by having relaxed the separability assumption. Figure \ref{fig:model-comparison} furthermore shows that the performance of GCFN depends on the choice of the decoder structure.

\subsection{High-dimensional Outcome}
\label{sec:high-dimensional-outcome}

In this section, we illustrate how Flow IV can be used to perform counterfactual inference for high-dimensional outcome variable $\Y$ as described in Section~\ref{sec:method-flow-matching}. This can be useful for tasks like counterfactual image editing \citep{PanBareinboim2024} under hidden confounding. Specifically, we perform counterfactual image editing on the MNIST dataset \citep{LeCun1998}.

We consider the following \ac{DGP}.

\begin{align*}
    &\text{DGP:} \quad Z \sim \Unif (0, 1), \qquad\qquad A = (Z + 0.2 \cdot U_a + \frac{(D + 1)}{10}) \frac{1}{2.2}, \qquad\qquad \Y = A \cdot \Ynoise\\
    &\text{with } D \sim \Unif\{0, 1\dots, 9\},\ U_a \sim \Unif (0, 1),\ \Ynoise \sim \Unif\{\text{MNIST dataset}\mid \text{Digit} = D\}
\end{align*}

The outcome variable is an MNIST image scaled by the treatment variable $A$ which takes values on $(0, 1)$. This implies that the triangular monotonicity assumption \ref{ass:triangular-monotonic} is satisfied as each output component $Y^{(i)}$ is only a function of $A$ and $U_{\y}^{(i)}$.
The confounder is the digit $D$ shown on the outcome sample, where $A$ tends to be larger for larger values of $D$. Therefore, higher digits tend to be multiplied by values of $A$ close to one and are therefore similar to the original MNIST images, whereas lower digits tend to be multiplied by values of $A$ close to zero and are therefore much dimmer than the original MNIST images. We compare the Flow IV method using flow matching to a purely associational model. We do not enforce the triangular structure of the true outcome function in our flow matching model, thereby violating the assumption made in Theorem \ref{thrm:identifiabiliy} (more detail in Appendix~\ref{app:trianguar-flow-matching}). The experiment serves as an ablation study to illustrate the effect of not enforcing the triangular structure. We compare the Flow IV model to an associational model, which also uses flow matching with the same architecture as in the Flow IV model but is only optimized to generate $\Y \mid A=a$ without utilizing the instrument $Z$.

\begin{figure}[ht]
    \centering
    \includegraphics[width=0.95\linewidth]{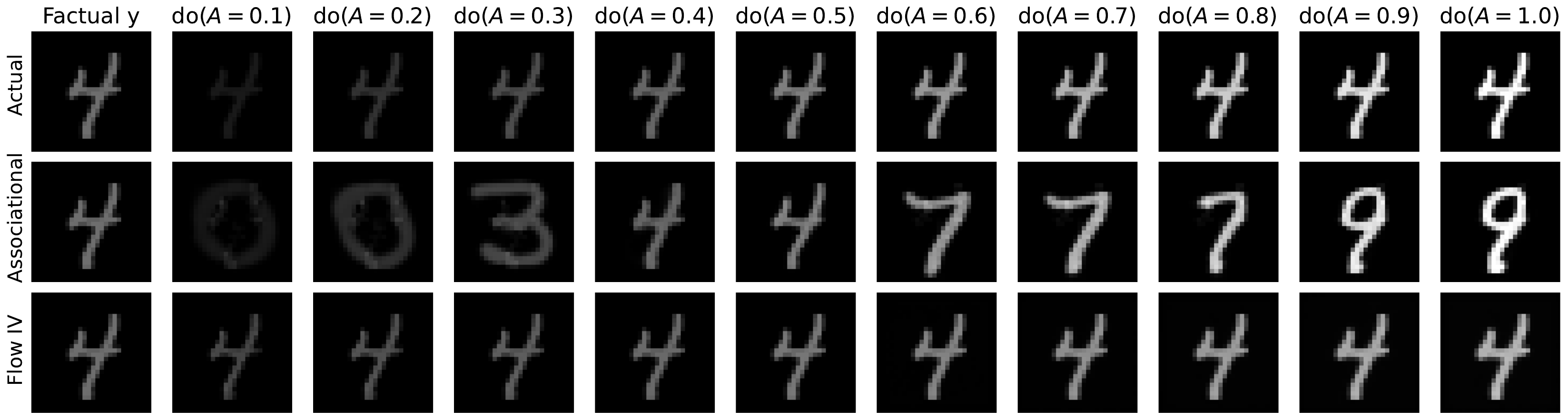}
    \caption{Comparison of counterfactual predictions for image data generated with Flow IV and a flow matching based associational model.}
    \label{fig:mnist-example}
\end{figure}

Figure \ref{fig:mnist-example} shows that the associational model seems to replicate the spurious association through the confounder (i.e. the digit shown in the image). When increasing the value of $A$, the associational model produces images of increasing digits, starting from digit zero at counterfactual level $A=0.1$ up to digit nine at counterfactual level $A=1.0$. Despite the fact that the triangular structure is not enforced, the Flow IV model correctly produces the digit four that can be seen on the factual image at all tested counterfactual levels of $A$. It seems like it slightly underestimates the effect of $A$ as the dimming effect is less pronounced compared to the ground truth counterfactual images. The better performance is quantitively confirmed by a cf-MSE of the Flow IV model of $2.025 \cdot 10^{-3}$ compared to $14.919 \cdot 10^{-3}$ of the associational flow matching model.

\subsection{Real-world data: The Impact of Chinese Foreign Aid on Economic Development}

IVs are common in social sciences and we replicate a foundational study in development economics, which influenced many subsequent studies. Specifically, we reanalyse an article investigating the impact of Chinese foreign aid on the economic development of different regions in Africa \citep{Dreher2021}. The article focuses on the role of political influence on the effectiveness of the supplied aid.
In a first step, the authors estimate the causal effect of Chinese foreign aid in USD per capita provided in the previous year (treatment $A$) on the nighttime light emission of that region which is used as a proxy for economic development (outcome $Y$). Both $A$ and $Y$ are in log-scale. The nighttime light data used in the study comes from a dataset that was collected using weather satellites from the US Air Force \citep{noaa2014} and the data on Chinese foreign aid was introduced by \citeauthor{Dreher2019}. The study uses the Chinese steel production as an instrument $Z$ (data from \citeauthor{WSA2010} \citeyear{WSA2010, WSA2014}).

The authors of the original article perform their analysis with the Usual IV approach and control for fixed effects capturing the yearly production volumes of steel and the recipient region's probability of receiving aid. We replicate this approach but use the Flow IV with normalizing flows to model the structural equations.

\begin{figure}[ht]
    \centering

    \subfigure[$\E \lbrack Y \mid do(A=a)\rbrack$ under different values of $a$.]{
        \includegraphics[width=0.49\linewidth]{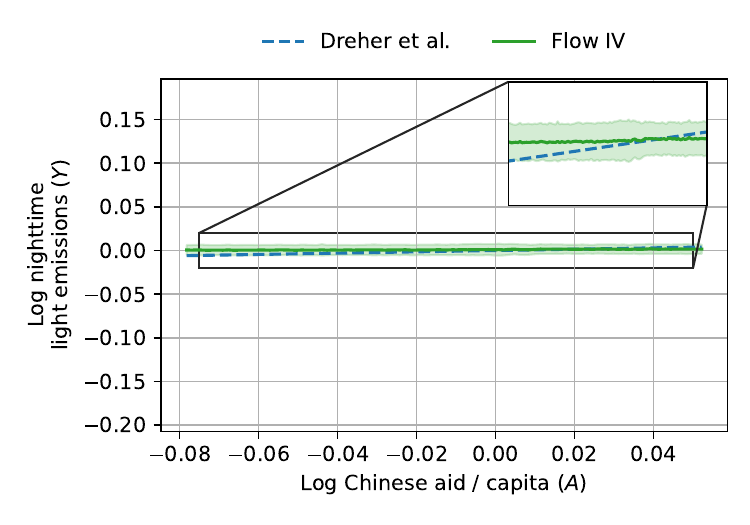}
        \label{fig:dreher-ace}
    }\hfill
    \subfigure[Outcome function learned by the Flow IV model. $q_i$ represents the $i$-quantile of $U_y$.]{
        \includegraphics[width=0.45\linewidth]{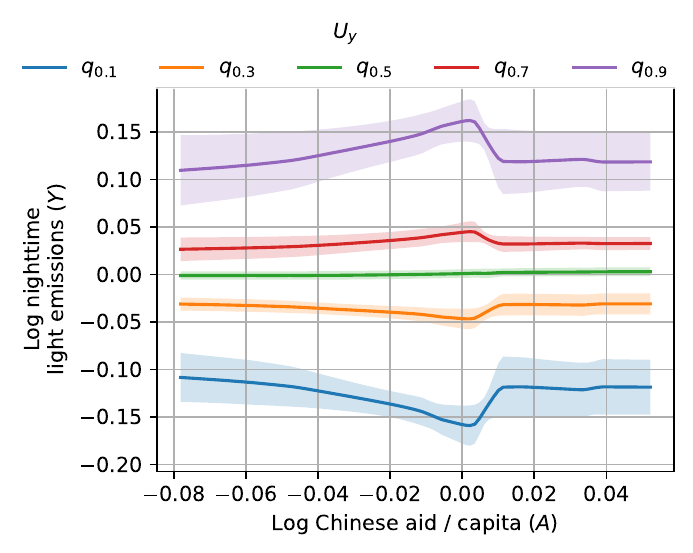}
        \label{fig:dreher-outcome-func}
    }

    \caption{The outcome function learned with the Flow IV model in (b) and the expectation over $U_y$ in (a).}
    \label{fig:dreher-combined}
\end{figure}

\citet{Dreher2021} found that the regression coefficient of the treatment variable was $0.0753$ (Table 1 column 1 in \citet{Dreher2021}) which corresponds to the constant treatment effect according to their model. Figure \ref{fig:dreher-ace} shows $\E \lbrack Y \mid do(A)\rbrack$ obtained from the Flow IV and the Usual IV estimate by \citet{Dreher2021}. According to the Flow IV model, we cannot conclude that the effect of $A$ on $Y$ is non-zero, whereas \citet{Dreher2021} find a small but significant positive effect. However, the estimates of \citet{Dreher2021} lie within the bootstrap confidence interval of the Flow IV method for most values of $A$, suggesting that the Flow IV model does not differ significantly from the linear model.

Since Flow IV aims at performing counterfactual inference, the method allows us to go beyond an analysis of average effects among all analysed African regions and to investigate the effect of Chinese foreign aid provided to specific individual regions.
Specifically, the learned outcome functions of Flow IV illustrated in Figure \ref{fig:dreher-outcome-func} can be used to assess counterfactuals according to the model. The abduction step corresponds to finding the (constant $U_y$) line which an observed pair $(y, a)$ lies on. For example, if we want to perform counterfactual inference on a region with $A=-0.06$ and $Y=0.12$, we can see in Figure \ref{fig:dreher-outcome-func} that the only value of $U_y$ that can produce this observation corresponds to the purple curve.
The action step corresponds to moving on the purple line (i.e. the value of $U_y$ is unchanged) from the observed value of $A=a$ to the counterfactual value $a'$. For example, if we want to know what the $Y$ value of the aforementioned region would have been had $A$ been equal to zero, then moving on the purple line to $A=0$ yields $Y'(a=0 \mid A=-0.06, Y=0.12) = 0.16$. Counterfactual inference allows us to make a personalized decision for this specific region: It would have seen the largest value in night-time light emissions (maximum of purple curve) for $A=0$.

In general, Figure \ref{fig:dreher-outcome-func} suggests that regions with high levels of nighttime light emissions (e.g. red and purple curves) might benefit from Chinese foreign aid up to a certain level (up to $A=0$, nighttime light emissions increase as the foreign aid increases) whereas countries with lower levels of nighttime light emissions might be harmed (orange and blue curve). This sort of counterfactual analysis can help decision-makers make more personalized interventions that account for regional heterogeneity in aid effectiveness.

\section{Discussion}
\label{sec:discussion}

We proposed Flow IV, a method that allows to perform counterfactual inference in nonseparable and high-dimensional outcome settings. In addition to common \ac{IV} assumptions, Flow IV relies on the assumption that the outcome function is triangular monotonic with respect to the latent factors, which for 1D outcomes is equivalent to invertibility. Invertibility is a necessary and sufficient condition for unique counterfactual inference. We showed that any correctly specified triangular monotonic outcome function that produces the observed distributions and that respects the independencies required by the \ac{IV} setup is consistent with the true counterfactual.
Therefore, any triangular monotonic generative model like autoregressive normalizing flows can make use of this result with Flow IV if the triangular ordering is known. We illustrated empirically how Flow IV with flow matching allows counterfactual image editing in the presence of hidden confounders even without encoding the triangular structure. How triangular monotonicity can be encoded in the structure of different network architectures in the flow matching approach and how $(\Ahatnoise, \Yhatnoise)$ can be modelled with flow matching instead of normalizing flows was not addressed in this paper and is thus an open question for future research.


\bibliography{references}

@Article{Imbens1994,
  author          = {Guido W. Imbens and Joshua D. Angrist},
  journal         = {Econometrica},
  title           = {Identification and Estimation of Local Average Treatment Effects},
  year            = {1994},
  issn            = {00129682, 14680262},
  number          = {2},
  pages           = {467--475},
  volume          = {62},
  comment-marbr77 = {Only consider Local Average Treatment Effect (LATE), i.e. the ATE conditioned on the subpopulation that complies with the instrument. I think they require monotonicity (?)},
  publisher       = {[Wiley, Econometric Society]},
  url             = {http://www.jstor.org/stable/2951620},
  urldate         = {2025-05-26},
}

@Book{Hernan2025,
  author          = {Hernán, Miguel A. and Robins, James M.},
  publisher       = {CRC Press},
  title           = {Causal Inference: What If},
  year            = {2025},
  comment-marbr77 = {Book on causality in general. Good overview of IVs. Propose some general conditions for identifiability, but can't find sources for some of them. Cov(e(U), t(U)) = 0 is generalisation of Wang and Tchetgen Tchetgen 2018 and there is a proof for it, but no source, so not sure if it is new.},
  url             = {https://miguelhernan.org/whatifbook},
}

@Article{Imbens2010,
  author          = {Guido W. Imbens},
  journal         = {Journal of Economic Literature},
  title           = {Better LATE Than Nothing: Some Comments on Deaton (2009) and Heckman and Urzua (2009)},
  year            = {2010},
  issn            = {00220515},
  number          = {2},
  pages           = {399--423},
  volume          = {48},
  comment-marbr77 = {Discusses why insights from observational studies are useful and why it is better to have a more precise estimate for a subpopulation than a inprecice estimate for the whole population.},
  publisher       = {American Economic Association},
  url             = {http://www.jstor.org/stable/20778730},
  urldate         = {2025-06-02},
}

@Article{Angrist1996,
  author    = {Joshua D. Angrist and Guido W. Imbens and Donald B. Rubin},
  journal   = {Journal of the American Statistical Association},
  title     = {Identification of Causal Effects Using Instrumental Variables},
  year      = {1996},
  issn      = {01621459},
  number    = {434},
  pages     = {444--455},
  volume    = {91},
  publisher = {[American Statistical Association, Taylor & Francis, Ltd.]},
  url       = {http://www.jstor.org/stable/2291629},
  urldate   = {2025-07-15},
}

@Article{Imbens2009,
  author            = {Guido W. Imbens and Whitney K. Newey},
  journal           = {Econometrica},
  title             = {Identification and Estimation of Triangular Simultaneous Equations Models without Additivity},
  year              = {2009},
  issn              = {00129682, 14680262},
  number            = {5},
  pages             = {1481--1512},
  volume            = {77},
  comment-marcbraun = {General Identification in the nonlinear continuous case. Consider Gaussian noise as special case. Do not estimate individual effects. Don't fully understand the paper yet.},
  publisher         = {[Wiley, The Econometric Society]},
  url               = {e},
  urldate           = {2025-07-09},
}

@Article{Chesher2003,
  author            = {Chesher, Andrew},
  journal           = {Econometrica},
  title             = {Identification in Nonseparable Models},
  year              = {2003},
  month             = {02},
  pages             = {1405-1441},
  volume            = {71},
  comment-marcbraun = {Identification for 1d output. Similar to our approach but fully non-parametric. They use quantile regression to estimate local derivatives dY/dA. Do not revocer full derivative / not the full structural equation Y = h(A), hence the term local identifiaciton.},
  doi               = {10.1111/1468-0262.00454},
}

@Article{Hahn2017,
  author            = {Jinyong Hahn and Geert Ridder},
  journal           = {Journal of Econometrics},
  title             = {Instrumental variable estimation of nonlinear models with nonclassical measurement error using control variables},
  year              = {2017},
  issn              = {0304-4076},
  note              = {Measurement Error Models},
  number            = {2},
  pages             = {238-250},
  volume            = {200},
  comment-marcbraun = {Assume additivity but don't fully understand this one yet.},
  doi               = {https://doi.org/10.1016/j.jeconom.2017.06.008},
  url               = {https://www.sciencedirect.com/science/article/pii/S0304407617300945},
}

@InProceedings{Hartford2017,
  author            = {Hartford, Jason and Lewis, Greg and Leyton-Brown, Kevin and Taddy, Matt},
  booktitle         = {Proceedings of the 34th International Conference on Machine Learning - Volume 70},
  title             = {Deep IV: a flexible approach for counterfactual prediction},
  year              = {2017},
  address           = {Sydney, NSW, Australia},
  pages             = {1414–1423},
  publisher         = {JMLR.org},
  series            = {ICML'17},
  comment-marcbraun = {Use neural networks but assume additive noise},
  numpages          = {10},
}

@InProceedings{Singh2019,
  author            = {Singh, Rahul and Sahani, Maneesh and Gretton, Arthur},
  booktitle         = {Advances in Neural Information Processing Systems},
  title             = {Kernel Instrumental Variable Regression},
  year              = {2019},
  editor            = {H. Wallach and H. Larochelle and A. Beygelzimer and F. d\textquotesingle Alch\'{e}-Buc and E. Fox and R. Garnett},
  publisher         = {Curran Associates, Inc.},
  volume            = {32},
  comment-marcbraun = {Assume additive noise},
  url               = {https://proceedings.neurips.cc/paper_files/paper/2019/file/17b3c7061788dbe82de5abe9f6fe22b3-Paper.pdf},
}

@Article{Guo2016,
  author  = {Zijian Guo and Dylan S. Small},
  journal = {Journal of Machine Learning Research},
  title   = {Control Function Instrumental Variable Estimation of Nonlinear Causal Effect Models},
  year    = {2016},
  number  = {100},
  pages   = {1--35},
  volume  = {17},
  url     = {http://jmlr.org/papers/v17/14-379.html},
}

@InProceedings{Puli2020,
  author            = {Puli, Aahlad and Ranganath, Rajesh},
  booktitle         = {Advances in Neural Information Processing Systems},
  title             = {General Control Functions for Causal Effect Estimation from IVs},
  year              = {2020},
  editor            = {H. Larochelle and M. Ranzato and R. Hadsell and M.F. Balcan and H. Lin},
  pages             = {8440--8451},
  publisher         = {Curran Associates, Inc.},
  volume            = {33},
  comment-marcbraun = {Very general, works for continuous, discrete, don't require linearity or separability. Only disadvantage is that they cannot identify counterfactuals, but they have much weaker assumptions than us in general. They also mention the Imbens and Newey 2009 paper and restate that under their monotonicity assumption of the confounder, their method also aways works.},
  url               = {https://proceedings.neurips.cc/paper_files/paper/2020/file/604f2c31e67034642b288d76a8df11d5-Paper.pdf},
}

@Book{Pearl2009,
  author    = {Pearl, Judea},
  publisher = {Cambridge University Press},
  title     = {Causality: Models, Reasoning and Inference},
  year      = {2009},
  address   = {USA},
  url       = {https://dl.acm.org/doi/book/10.5555/1642718},
}

@Article{Wald1940,
  author    = {Abraham Wald},
  journal   = {The Annals of Mathematical Statistics},
  title     = {{The Fitting of Straight Lines if Both Variables are Subject to Error}},
  year      = {1940},
  number    = {3},
  pages     = {284 -- 300},
  volume    = {11},
  doi       = {10.1214/aoms/1177731868},
  publisher = {Institute of Mathematical Statistics},
  url       = {https://doi.org/10.1214/aoms/1177731868},
}

@Misc{Liu2025,
  author        = {Jiewen Liu and Chan Park and Yonghoon Lee and Yunshu Zhang and Mengxin Yu and James M. Robins and Eric J. Tchetgen Tchetgen},
  title         = {The Multiplicative Instrumental Variable Model},
  year          = {2025},
  archiveprefix = {arXiv},
  eprint        = {2507.09302},
  primaryclass  = {stat.ME},
  url           = {https://arxiv.org/abs/2507.09302},
}

@Article{Robins1994,
  author    = {James M. Robins},
  journal   = {Communications in Statistics - Theory and Methods},
  title     = {Correcting for non-compliance in randomized trials using structural nested mean models},
  year      = {1994},
  number    = {8},
  pages     = {2379--2412},
  volume    = {23},
  doi       = {10.1080/03610929408831393},
  publisher = {Taylor \& Francis},
  url       = {https://doi.org/10.1080/03610929408831393},
}

@InProceedings{balgi2022,
  author    = {Sourabh Balgi and Adel Daoud and Jos{\'{e}} M. Pe{\~{n}}a},
  booktitle = {Proceedings of the AAAI Conference on Artificial Intelligence},
  title     = {Personalized public policy analysis in social sciences using causal-graphical normalizing flows},
  year      = {2022},
  pages     = {11810--11818},
  volume    = {36},
  doi       = {10.1609/aaai.v36i11.21437},
  url       = {https://cdn.aaai.org/ojs/21437/21437-13-25450-1-2-20220628.pdf},
}

@Article{Balgi2024,
  author   = {Sourabh Balgi and Adel Daoud and Jos{\'{e}} M. Pe{\~{n}}a and Geoffrey Wodtke and Jesse Zhou},
  journal  = {Sociological Methods \& Research},
  title    = {Deep Learning With DAGs},
  year     = {2024},
  doi      = {10.1177/00491241251319291},
  eprint   = {https://doi.org/10.1177/00491241251319291},
  url      = {https://doi.org/10.1177/00491241251319291},
}

@InProceedings{Javaloy2023,
  author    = {Javaloy, Adri\'{a}n and Sanchez-Martin, Pablo and Valera, Isabel},
  booktitle = {Advances in Neural Information Processing Systems},
  title     = {Causal normalizing flows: from theory to practice},
  year      = {2023},
  editor    = {A. Oh and T. Naumann and A. Globerson and K. Saenko and M. Hardt and S. Levine},
  pages     = {58833--58864},
  publisher = {Curran Associates, Inc.},
  volume    = {36},
  url       = {https://proceedings.neurips.cc/paper_files/paper/2023/file/b8402301e7f06bdc97a31bfaa653dc32-Paper-Conference.pdf},
}

@InProceedings{Kingma2016,
  author    = {Kingma, Durk P and Salimans, Tim and Jozefowicz, Rafal and Chen, Xi and Sutskever, Ilya and Welling, Max},
  booktitle = {Advances in Neural Information Processing Systems},
  title     = {Improved Variational Inference with Inverse Autoregressive Flow},
  year      = {2016},
  editor    = {D. Lee and M. Sugiyama and U. Luxburg and I. Guyon and R. Garnett},
  publisher = {Curran Associates, Inc.},
  volume    = {29},
  url       = {https://proceedings.neurips.cc/paper_files/paper/2016/file/ddeebdeefdb7e7e7a697e1c3e3d8ef54-Paper.pdf},
}

@Article{Dreher2021,
  author   = {Axel Dreher and Andreas Fuchs and Roland Hodler and Bradley C. Parks and Paul A. Raschky and Michael J. Tierney},
  journal  = {World Development},
  title    = {Is Favoritism a Threat to Chinese Aid Effectiveness? A Subnational Analysis of Chinese Development Projects},
  year     = {2021},
  issn     = {0305-750X},
  pages    = {105291},
  volume   = {139},
  doi      = {https://doi.org/10.1016/j.worlddev.2020.105291},
  url      = {https://www.sciencedirect.com/science/article/pii/S0305750X20304186},
}

@Misc{noaa2014,
  author       = {NOAA},
  howpublished = {https://eogdata.mines.edu/products/dmsp/},
  note         = {Accessed: 2014-04-05},
  title        = {National Oceanic and Atomspheric Administration , Version 4 DMSP-OLS Nighttime Lights Time Series. Boulder, CO: National Geophysical Data Center},
  year         = {2014},
}

@Article{Dreher2019,
  author   = {Axel Dreher and Andreas Fuchs and Roland Hodler and Bradley C. Parks and Paul A. Raschky and Michael J. Tierney},
  journal  = {Journal of Development Economics},
  title    = {African leaders and the geography of China's foreign assistance},
  year     = {2019},
  issn     = {0304-3878},
  pages    = {44-71},
  volume   = {140},
  doi      = {https://doi.org/10.1016/j.jdeveco.2019.04.003},
  url      = {https://www.sciencedirect.com/science/article/pii/S030438781831099X},
}

@Misc{WSA2010,
  author       = {WSA},
  howpublished = {https://worldsteel.org/wp-content/uploads/Steel-Statistical-Yearbook-2010.pdf},
  title        = {World Steel Association, Statistical Yearbook 2010, Brussels, Belgium: Worldsteel Committee on Economic Studies},
  year         = {2010},
}

@Misc{WSA2014,
  author       = {WSA},
  howpublished = {https://worldsteel.org/wp-content/uploads/Steel-Statistical-Yearbook-2014.pdf},
  title        = {World Steel Association, Statistical Yearbook 2014, Brussels, Belgium: Worldsteel Committee on Economic Studies},
  year         = {2014},
}

@Article{Pearl1995,
  author    = {Judea Pearl},
  journal   = {Biometrika},
  title     = {Causal Diagrams for Empirical Research},
  year      = {1995},
  issn      = {00063444, 14643510},
  number    = {4},
  pages     = {669--688},
  volume    = {82},
  publisher = {[Oxford University Press, Biometrika Trust]},
  url       = {http://www.jstor.org/stable/2337329},
  urldate   = {2025-07-30},
}

@Article{Lin2023,
  author  = {Lin, Cheng and Peña, Jos{\'{e}} M. and Daoud, Adel},
  journal = {arXiv preprint arXiv:2311.13410},
  title   = {Assessing the Unobserved: Enhancing Causal Inference in Sociology with Sensitivity Analysis},
  year    = {2023},
}

@InProceedings{Lipman2023,
  author    = {Yaron Lipman and Ricky T. Q. Chen and Heli Ben-Hamu and Maximilian Nickel and Matthew Le},
  booktitle = {The Eleventh International Conference on Learning Representations},
  title     = {Flow Matching for Generative Modeling},
  year      = {2023},
  url       = {https://openreview.net/forum?id=PqvMRDCJT9t},
}

@Article{Ogburn2012,
  author    = {Ogburn, Elizabeth L. and VanderWeele, Tyler J.},
  journal   = {Epidemiology},
  title     = {On the Nondifferential Misclassification of a Binary Confounder},
  year      = {2012},
  issn      = {1044-3983},
  month     = may,
  number    = {3},
  pages     = {433--439},
  volume    = {23},
  doi       = {10.1097/ede.0b013e31824d1f63},
  publisher = {Ovid Technologies (Wolters Kluwer Health)},
}

@Misc{LeCun1998,
  author       = {LeCun, Yann and Cortes, Corinna and Burges, Christopher J.C.},
  howpublished = {\url{http://yann.lecun.com/exdb/mnist/}},
  title        = {The MNIST database of handwritten digits},
  year         = {1998},
}

@Article{Kino2021,
  author   = {Shiho Kino and Yu-Tien Hsu and Koichiro Shiba and Yung-Shin Chien and Carol Mita and Ichiro Kawachi and Adel Daoud},
  journal  = {SSM - Population Health},
  title    = {A scoping review on the use of machine learning in research on social determinants of health: Trends and research prospects},
  year     = {2021},
  issn     = {2352-8273},
  pages    = {100836},
  volume   = {15},
  doi      = {https://doi.org/10.1016/j.ssmph.2021.100836},
  url      = {https://www.sciencedirect.com/science/article/pii/S2352827321001117},
}

@InProceedings{NasrEsfahany2023,
  author    = {Arash Nasr-Esfahany and MohammadIman Alizadeh and Devavrat Shah},
  booktitle = {International Conference on Machine Learning},
  title     = {Counterfactual Identifiability of Bijective Causal Models},
  year      = {2023},
}

@InProceedings{PanBareinboim2024,
  title = 	 {Counterfactual Image Editing},
  author =       {Pan, Yushu and Bareinboim, Elias},
  booktitle = 	 {Proceedings of the 41st International Conference on Machine Learning},
  pages = 	 {39087--39101},
  year = 	 {2024},
  editor = 	 {Salakhutdinov, Ruslan and Kolter, Zico and Heller, Katherine and Weller, Adrian and Oliver, Nuria and Scarlett, Jonathan and Berkenkamp, Felix},
  volume = 	 {235},
  series = 	 {Proceedings of Machine Learning Research},
  month = 	 {21--27 Jul},
  publisher =    {PMLR},
  url = 	 {https://proceedings.mlr.press/v235/pan24a.html}
}

@Article{Papamakarios2021,
  author     = {Papamakarios, George and Nalisnick, Eric and Rezende, Danilo Jimenez and Mohamed, Shakir and Lakshminarayanan, Balaji},
  journal    = {J. Mach. Learn. Res.},
  title      = {Normalizing flows for probabilistic modeling and inference},
  year       = {2021},
  issn       = {1532-4435},
  month      = jan,
  number     = {1},
  volume     = {22},
  articleno  = {57},
  issue_date = {January 2021},
  numpages   = {64},
  publisher  = {JMLR.org},
}

\appendix

\appendix
%
%
%
%
\section{Technical Proofs}
\label{app:technical-proofs}

We start by proving the following claim made in Section~\ref{sec:problem-setup}.

\begin{claim}[Justification of Assumption \ref{ass:triangular-monotonic}]
\label{claim:invertibility}
When the necessary and sufficient condition for unique counterfactual inference stated in Equation~\ref{cond:unique-cf} is true, then we can rewrite \ac{SCM} \ref{scm:setup} such that the outcome function is invertible while the model stays counterfactually equivalent.
\end{claim}

\begin{example}
    Assume the outcome function is $f(A, U_y) = A + U_y^2$. Then we can only determine $u_y$ in the abduction step up to its sign. However, note that all $\pm u_y$ result in the same counterfactual outcome under all alternative treatments (i.e. Condition \ref{cond:unique-cf} is true). We can therefore just replace $f(A, U_y)$ with $f^\star(A, U_y) = A + U^\star_y$ for $U^\star_y = U_y^2$ where then $f$ and $f^\star$ are counterfactually equivalent.
\end{example}

\begin{proof}
    The equivalency in Equation~\ref{cond:unique-cf} states the following necessary condition for unique counterfactual inference.
    \begin{align*}
        f_{\y}(\a, {\ynoise}) = f_{\y}(\a, {\ynoise}') \iff f_{\y}(\a', {\ynoise}) = f_{\y}(\a', {\ynoise}')
    \end{align*}
    Assume this condition is true. Now let $\Tilde{f}_{\y}: \mathcal{X} \rightarrow \Ydomain$ be a function with domain $\mathcal{X} \subseteq \Adomain \times \Ynoisedomain$ such that $\Tilde{f}_{\y}(\a, {\ynoise}) = f_{\y}(\a, {\ynoise})$ for all $(\a, \ynoise) \in \mathcal{X}$. Let $\mathcal{X}$ contain all elements of $\Adomain \times \Ynoisedomain$ except for pairs $(\a, \ynoise)$ and $(\a, \ynoise')$ for which $f_{\y}(\a, {\ynoise}) = f_{\y}(\a, {\ynoise}')$. For those pairs, only keep $(\a,\ynoise)$. This ensures that $\Tilde{f}_{\y}$ is invertible.

    The counterfactual outcome under alternative treatment $\a'$ given the observations $\y$ and $\a$ with the model $f_{\y}$ is

    \begin{align*}
        \{\ynoise, \ynoise'\} &= f_{\y}^{-1}(\a, \y) \tag*{\text{(Abduction)}} \\
        \Y(\a') &= f_{\y}(\a', \ynoise) = f_{\y}(\a', \ynoise') \tag*{\text{(Action), (Cond. in Eq. \ref{cond:unique-cf})}}
    \end{align*}
    where $f_{\y}^{-1}$ is the multivalued inverse of $f_{\y}$.

    The same counterfactual with the model $\Tilde{f}_{\y}$ yields
    \begin{align*}
        \ynoise &= \Tilde{f}_{\y}^{-1}(\a, \y) \tag*{\text{(Abduction), (Def. of $\Tilde{f}_{\y}$)}} \\
        \Tilde{\Y}(\a') &= \Tilde{f}_{\y}(\a', \ynoise) \tag*{\text{(Action)}} \\
        &= f_{\y}(\a', \ynoise) \tag*{\text{(Def. of $\Tilde{f}_{\y}$)}} \\
        &= \Y(\a')
    \end{align*}
    which concludes the proof.
\end{proof}

Before we prove Theorem \ref{thrm:identifiabiliy}, we will prove the following lemma.

\begin{lemma}
\label{lemma:triangular-monotonic}
    Let $f_{\alpha}: \R^n \rightarrow \R^n$ be a function with parameter $\alpha$ and $g_{\beta}: \R^n \rightarrow \R^n$ be a function with parameter $\beta$. Let $f_{\alpha}$ and $g_{\beta}$ be triangular monotonic as defined in Assumption~\ref{ass:triangular-monotonic}. Let $X$ be a continuous random variable and define $\tau \coloneq f_{\alpha}(X)$ and $\eta \coloneq g_{\beta}(X)$.
    For any choice of $\alpha$ and $\beta$ let
    \begin{equation*}
        f_{\alpha}(X) \overset{d}{=} g_{\beta}(X)
    \end{equation*}
    It then follows that
    \begin{equation*}
        f_{\alpha}(x) = g_{\beta}(x) \qquad \text{a.s.}
    \end{equation*}
\end{lemma}

\begin{proof}
    By assumption we have
    \begin{equation*}
        \tau \overset{d}{=} \eta
    \end{equation*}
    which implies that
    \begin{align*}
        \tau^{(1)} \overset{d}&{=} \eta^{(1)} \\
        \tau^{(2)} \mid \tau^{(1)} \overset{d}&{=} \eta^{(2)} \mid \eta^{(1)} \\
        &\hdots \\
        \tau^{(n)} \mid \tau^{(n-1)} \dots \tau^{(1)} \overset{d}&{=} \eta^{(n)} \mid \eta^{(n-1)} \dots \eta^{(1)}
    \end{align*}

    However, because each $(f_{\alpha}^{(i)}, f_{\alpha}^{(i-1)}, \dots f_{\alpha}^{(1)})$ is a bijective mapping between $(\tau^{(i)}, \tau^{(i-1)}, \dots, \tau^{(1)})$ and $(X^{(i)}, X^{(i-1)}, \dots, X^{(1)})$ due to the strict monotonicity assumption, we can write each $\tau^{(i)} \mid \tau^{(i-1)}, \dots, \tau^{1}$ as $\tau^{(i)} \mid x^{(i-1)}, \dots, x^{1}$. The same is true for $\eta$ and $g_{\beta}$. We then have

    \begin{align*}
        \tau^{(i)} \mid x^{(i-1)}, \dots, x^{1} \overset{d}&{=} \eta^{(i)} \mid x^{(i-1)}, \dots, x^{1} \\
        \iff f_{\alpha}^{(i)}(X^{(i)}, x^{(i-1)}, \dots, x^{(1)}) \overset{d}&{=}  g_{\beta}^{(i)}(X^{(i)}, x^{(i-1)}, \dots, x^{(1)}) \\
        \iff f_{\alpha}^{(i)}(x^{(i)}, x^{(i-1)}, \dots, x^{(1)}) &=  g_{\beta}^{(i)}(x^{(i)}, x^{(i-1)}, \dots, x^{(1)}) \qquad \text{a.s.}
    \end{align*}

    where the last equality follows from the strict monotonicity of $f_{\alpha}^{(i)}$ and $g_{\beta}^{(i)}$ w.r.t. $X^{(i)}$.
\end{proof}

Next, we show the proof of Theorem \ref{thrm:identifiabiliy}.

\begin{proof} (\textbf{Theorem} \ref{thrm:identifiabiliy})

Let $\Z$, $\A$, and $\Y$ be random variables with nonzero continuous densities in $\Zdomain \subseteq \R^k$, $\Adomain \subseteq \R^m$, and $\Ydomain \subseteq \R^n$ respectively. Let $\Ynoise$, and $\Yhatnoise$ be unobserved random variables with nonzero continuous densities in $\Ynoisedomain \subseteq \R^n$ and let $\Anoise$ be an unobserved random variable with nonzero continuous density in $\Anoisedomain \subseteq \R^w$.
Given the independencies specified by the graph in Figure \ref{fig:iv-dag}, assume that the output $\Y$ is generated by an unknown function $f_{\y}: \Adomain \times \Ynoisedomain \rightarrow \Ydomain$ such that $\Y \coloneq f_{\y}(\A, \Ynoise)$. We observe the conditional distribution of $\Y \mid \Z = \z, \A = \a$. We want to find a function $g_{\y}: \R^m \times \R^n \rightarrow \R^n$ such that $\Ycond \overset{d}{=} g(\a, \Yhatnoisecond)$ where $\Yhatnoise$ has the same independencies as $\Ynoise$ and $\overset{d}{=}$ denotes equality in distribution.

By Assumption \ref{ass:triangular-monotonic}, $f_{\y}$ is invertible and we have

\begin{align}
    \Ynoisecond \overset{d}&{=} f_{\y}^{-1}(\a, g_{\y}(\a, \Yhatnoisecond))
\end{align}

where $f_{\y}^{-1}(\cdot, \ynoise)$ denotes the inverse of $f_{\y}(\cdot, \ynoise)$.

Let $f_{\a}^{-1}: \Zdomain \times \Adomain \rightrightarrows \Anoisedomain$ be the possibly multivalued inverse of the treatment function such that $f_{\a}^{-1}(\z, \a) = \{\anoise \in \Adomain \mid \a = f_{\a}(\z, \anoise) \}$ where $f_{\a}$ is the treatment function.

Let $\Ynoise^c$ be a random variable with the same distribution as $\Ynoise \mid \Anoise \in f_{\a}^{-1}(\z, \a)$ and let $\Yhatnoise^c$ be similarly defined for $\Yhatnoise \mid \Anoise \in f_{\a}^{-1}(\z, \a)$. We then have

\begin{align}
    \Ynoise^c
    \label{eq:f-equality}
    \overset{d}&{=} f_{\y}^{-1}\left(\ \a,\ g\left( \a,\ (\Yhatnoise^c \right) \ \right)
\end{align}

Let $\bm{x}^{(1:i)} \coloneq (x^{(1)}, \dots, x^{(i)})^T$ and $\bm{x}^{(1:0)} = \emptyset$ and define

\begin{align}
    q(\ynoise; \anoise) &\coloneq (q^{(1)}(\ynoiseuniv^{(1)}; \anoise, \ynoise^{(1:0)}), q^{(2)}(\ynoiseuniv^{(2)}; \anoise, \ynoise^{(1:1)}), \dots, q^{(n)}(\ynoiseuniv^{(n)}; \anoise, \ynoise^{(1:n-1)}))^T \\
    r(\yhatnoise; \anoise) &\coloneq (r^{(1)}(\yhatnoiseuniv^{(1)}; \anoise, \yhatnoise^{(1:0)}), r^{(2)}(\yhatnoiseuniv^{(2)}; \anoise, \yhatnoise^{(1:1)}), \dots, r^{(n)}(\yhatnoiseuniv^{(n)}; \anoise, \yhatnoise^{(1:n-1)}))^T
\end{align}

as the vectors of unobserved (univariate) conditional cumulative distribution functions (CDFs). This means each component is defined as

\begin{align}
    q^{(i)}(u_{\y}^{(i)}; \anoise, \ynoise^{(1:i-1)}) &\coloneq \Pr(U_{\y}^{(i)} \leq u_{\y}{(i)} \mid \Anoise \in \anoise, \Ynoise^{(1:i-1)} = \ynoise^{(1:i-1)}) \\
    r(\yhatnoise^{(i)}; \anoise, \yhatnoise^{(1:i-1)}) &\coloneq \Pr(\tilde{U}_{\y}^{(i)} \leq \yhatnoise{(i)} \mid \Anoise \in \anoise, \Yhatnoise^{(1:i-1)} = \yhatnoise^{(1:i-1)})
\end{align}

Let the inverse of $q(\ynoise; \anoise)$ be called $q^{-1}(u; \anoise)$.

We can then use the Rosenblatt transform with $q$ and $r$ to transform $(\Ynoise \mid \Anoise = \anoise)$ and $(\Yhatnoise \mid \Anoise = \anoise)$ respectively to a multivariate uniform random variable

\begin{align}
    q((\Ynoise \mid \Anoise \in \anoise); \anoise) \overset{d}&{=} r((\Yhatnoise \mid \Anoise \in \anoise); \anoise) \sim \Unif([0, 1]^n) \\
    \label{eq:cdf-equality}
    \iff (\Ynoise \mid \Anoise \in \anoise)
    \overset{d}&{=} q^{-1}(r((\Yhatnoise \mid \Anoise \in \anoise); \anoise); \anoise)
\end{align}

Together with Equation~\ref{eq:f-equality} this results in the following equality.

\begin{align}
    q^{-1}(r(\Yhatnoise^c; f_{\a}^{-1}(\z, \a)); f_{\a}^{-1}(\z, \a)) \overset{d}&{=} f_{\y}^{-1} \left( \a, g_{\y}(\a,( \Yhatnoise^c) \right)
\end{align}

Because the left-hand side (LHS) is triangular monotonic as a consequence of the Rosenblatt transform and the right-hand side (RHS) expression is triangular monotonic according to Assumption~\ref{ass:triangular-monotonic}, from Lemma~\ref{lemma:triangular-monotonic} it follows that that for all $\yhatnoise \in \{\yhatnoise \in \Ynoisedomain \mid \Pr(\Yhatnoise^c) > 0 \}$

\begin{align}
    q^{-1}(r(\yhatnoise; f_{\a}^{-1}(\z, \a)); f_{\a}^{-1}(\z, \a)) &= f_{\y}^{-1} \left( \a, g_{\y}(\a, \yhatnoise) \right)
\end{align}

Because $f_{\a}$ is a deterministic function and because $\A$ and $\Z$ are not independent according to the causal graph in Figure \ref{fig:iv-dag}, Assumption \ref{ass:strong-instrument} implies that for every tuple $(\a, \anoise)$ with $\a \in \Adomain,\ \anoise \in \Anoisedomain$ there exists $\z \in \Zdomain$ such that $\anoise \in f_{\a}^{-1}(\z, \a)$.

Then, we know that for all $\anoise \in \Anoisedomain$ we have that

\begin{align}
    q^{-1}(r(\yhatnoise; \anoise); \anoise) &= f_{\y}^{-1} \left( \a, g_{\y}(\a, \yhatnoise) \right)
\end{align}

Because the LHS does not depend on $\a$ and the RHS does not depend on $\anoise$, and because both sides are invertible w.r.t. $\yhatnoise$, we can conclude that there has to exist an invertible function $\psi: \Ydomain \rightarrow \Ydomain$ that only depends on $\yhatnoise$ such that

\begin{align}
    f_{\y}^{-1} \left( \a, g_{\y}(\a,\yhatnoise) \right) &= \psi(\yhatnoise) \\
    \iff g_{\y}(\a, \yhatnoise) &= f_{\y} \left(\a, \psi(\yhatnoise) \right)
\end{align}

We have shown that if Assumptions \ref{ass:triangular-monotonic} and \ref{ass:strong-instrument} are true, then all triangular monotonic $g_{\y}$ (following the same triangular ordering as $f_{\y}$) that produce the correct observed distribution differs from $f_{\y}$ only by some invertible transformation of the parameter $\yhatnoise$ which concludes the proof of Theorem \ref{thrm:identifiabiliy}.

\end{proof}

\begin{claim}[Justification of the implication after Theorem~\ref{thrm:identifiabiliy}]
\label{claim:cf-equivalence}
Under the conditions of Theorem~\ref{thrm:identifiabiliy}, using
\( g_{\y} \) in the abduction and prediction steps yields the same 
counterfactual value as using \( f_{\y} \).
\end{claim}

\begin{proof}
    Let $g_{\y}(\a, \yhatnoise) = f_{\y} \left(\a, \psi(\yhatnoise) \right)$.
    The counterfactual outcome under the model $\y = f_{\y}(\a, \ynoise)$ is
    \begin{align*}
        \ynoise &= f_{\y}^{-1}(\a, \y) \tag*{\text{(Abduction)}} \\
        \Y(\a') &= f_{\y}(\a', \ynoise) \tag*{\text{(Action)}} \\
        &= f_{\y}(\a', f_{\y}^{-1}(\a, \y))
    \end{align*}
    
    The counterfactual outcome under the model $\hat{\y} = g_{\y}(\a, \yhatnoise) = f_{\y}(\a, \psi(\yhatnoise))$ is
    \begin{align*}
        \yhatnoise &= g^{-1}(\a, \y) \tag*{\text{(Abduction)}} \\
        &= \psi^{-1}(f_{\y}^{-1}(\a, \y)) \\
        \hat{\Y}(\a') &= g_{\y}(\a', \yhatnoise) \tag*{\text{(Action)}} \\
        &= f_{\y}(\a', \psi(\yhatnoise)) \\
        &= f_{\y}(\a', \psi(\psi^{-1}(f_{\y}^{-1}(\a, \y))) \\
        &= f_{\y}(\a', f_{\y}^{-1}(\a, \y)) \\
        &= \Y(\a')
    \end{align*}
    which concludes the proof.
\end{proof}
%
%
%
%
\section{Experiments}
\label{app:experiments}

\subsection{Synthetic Data Generation}
\label{app:synthetic-data}

In Section~\ref{sec:experiments} we introduce three different \acp{DGP} that represent the assumptions that the different \ac{IV} approaches make. We specify the full \acp{DGP} here.

\begin{align*}
    \text{DGP 1:} \qquad  &Z = U_Z \\
    &A = Z + U_A \\
    &Y = 0.6 \cdot A + U_Y \\
    &U_Y = \left( \alpha \cdot U_A^2 + \frac{1}{8}\eta - \alpha \right) \left( 2 \alpha^2 + \frac{1}{64} \right)^{-\frac{1}{2}} \\
    &\text{with } U_A, \eta \overset{iid}{\sim} \mathcal{N}(0,1) \\[0.8em]
    \text{DGP 2:} \qquad &Z = U_Z \\
    &A = Z + \sin (U_A) \\
    &Y = \left(\sin{(A+1.5)} + 1 \right) \cdot U_Y\\
    &U_Y = \left( \alpha \cdot U_A^2 + \frac{1}{8}\eta - \alpha \right) \left( 2 \alpha^2 + \frac{1}{64} \right)^{-\frac{1}{2}} \\
    &\text{with } U_A, \eta \overset{iid}{\sim} \mathcal{N}(0,1) \\[0.8em]
    \text{DGP 3:} \qquad  &Z = U_Z \\
    &A = Z + U_A \\
    &Y = 0.6 \cdot (A + U_Y)^2 \\
    & (U_A, U_Y)^T \sim \mathcal{N}(\mathbf{0}, \begin{pmatrix}
        1 & \rho \\
        \rho & 1
    \end{pmatrix}) \\
    &\text{with } \rho = -\exp(-\alpha) + 1
\end{align*}

\subsection{Additional Experiments}

In addition to the experiments in Section~\ref{sec:experiments}, we performed ablation studies to investigate the predictive quality of Flow IV for counterfactuals when different assumptions are violated. We furthermore investigated the convergence of Flow IV under finite samples and the performance of Flow IV with flow matching when we enforce triangularity of the vector field network.

\subsubsection{Violation of Flow IV's Assumptions}

We present two experiments that investigate the performance of Flow IV when the two of its core assumptions are violated.

\paragraph{Wrong Triangular Structure}

One of the assumptions of Flow IV is, that the outcome function of the true underlying \ac{DGP} has a triangular structure and that the learned outcome function has the same triangular structure. We use the following \ac{DGP} to train two Flow IV models, one with the correct and one with the reversed triangular structure, and a purely associational model in the sense that it assumes unconfoundedness.

\begin{align*}
    Z &= U_Z \\
    A &= 0.5 (Z + U_A) \\
    Y_1 &= A + U_{Y_1} \\
    Y_2 &= (\left|A\right| + 0.1) U_{Y_1} + U_{Y_2} \\
    \Y &= \begin{pmatrix}
        Y_1 \\
        Y_2
    \end{pmatrix} \\
    \text{with}\qquad\qquad U_Z &\sim \Normal{0}{1} \\
    \begin{pmatrix}
        U_A \\
        U_{Y_1} \\
        U_{Y_2}
    \end{pmatrix}
    &\sim \Normal{\mathbf{0}}{\begin{pmatrix}
1 & \rho & \rho\\
\rho & 1 & \rho^2\\
\rho & \rho^2 & 1
\end{pmatrix}}, \quad \rho = 0.5 \\
\end{align*}

For the model Flow IV (correct triangularity) we assume that $Y_1$ is a function only of $U_{Y_1}$ and $Y_2$ is a function of $U_{Y_1}$ and $U_{Y_2}$ just like it is the case in the \ac{DGP}. For the model Flow IV (wrong triangularity) we assume the opposite, i.e. $Y_2$ is a function only of $U_{Y_1}$ and $Y_1$ is a function of $U_{Y_1}$ and $U_{Y_2}$. The results can be seen in Table~\ref{tab:wrong-triangularity-results}.

\begin{table}[ht]
\centering
\begin{tabular}{lccc}
\toprule
Model & \makecell{Flow IV \\ (correct triangularity)} 
      & \makecell{Flow IV \\ (wrong triangularity)} 
      & Associational Model \\
\midrule
cf-RMSE & $0.180 \pm 0.047$ & $0.388 \pm 0.014$ & $0.860 \pm 0.023$ \\
\bottomrule
\end{tabular}
\caption{Comparison of counterfactual RMSE across models assuming wrong triangular structure.}
\label{tab:wrong-triangularity-results}
\end{table}

We can see that in this example Flow IV outperforms the associational model that assumes unconfoundedness even if the wrong triangular structure is used. However, the results also suggest that choosing the correct triangular structure is (in general) a necessary assumption as the model with the correct triangular structure outperformed the model that assumed the wrong triangular structure.

\paragraph{Non-Triangular Structure}

We furthermore ran an experiment with the following \ac{DGP} where the triangularity assumption is violated. Note that the outcome function is invertible, so counterfactual inference is in principle possible.

\begin{align*}
    Z &= U_Z \\
    A &= 0.5 (Z + U_A) \\
    Y_1 &= (\left|A\right| + 0.1) U_{Y_2} + U_{Y_1} \\
    Y_2 &= (\left|A\right| + 0.1) U_{Y_1} + U_{Y_2} \\
    \Y &= \begin{pmatrix}
        Y_1 \\
        Y_2
    \end{pmatrix} \\
    \text{with}\qquad\qquad U_Z &\sim \Normal{0}{1} \\
    \begin{pmatrix}
        U_A \\
        U_{Y_1} \\
        U_{Y_2}
    \end{pmatrix}
    &\sim \Normal{\mathbf{0}}{\begin{pmatrix}
1 & \rho & \rho\\
\rho & 1 & \rho^2\\
\rho & \rho^2 & 1
\end{pmatrix}}, \quad \rho = 0.5 \\
\end{align*}

From the results in Table~\ref{tab:non-triangularity-results} we can see that Flow IV does not outperform a model that assumes unconfoundedness in this experiment where the true \ac{DGP} is not triangular, underlying the importance of the triangularity assumption.

From the experiments in this section it can be concluded that the triangularity assumption and the correct triangular specification of the Flow IV model are important to ensure that we can predict counterfactuals consistently. Otherwise, Flow IV may be outperformed even by purely associational models.

\begin{table}[ht]
\centering
\begin{tabular}{lcc}
\toprule
Model & Flow IV
      & Associational Model \\
\midrule
cf-RMSE & $0.594 \pm 0.099$ & $0.454 \pm 0.057$ \\
\bottomrule
\end{tabular}
\caption{Comparison of counterfactual RMSE when the \ac{DGP} is non-triangular but invertible.}
\label{tab:non-triangularity-results}
\end{table}

\subsubsection{Finite-Sample Behaviour}

\begin{figure}[ht]
    \centering
    \includegraphics[width=0.65\linewidth]{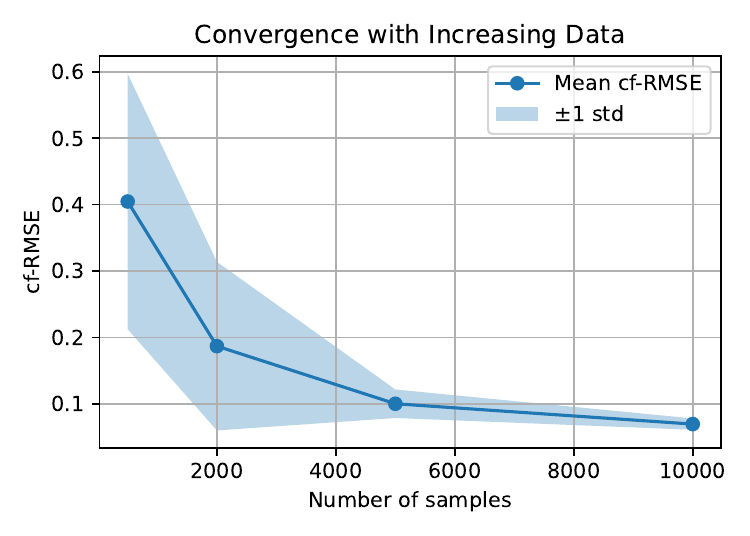}
    \caption{Finite sample convergence of Flow IV.}
    \label{fig:convergence}
\end{figure}

Using the following \ac{DGP}, we investigated the convergence of Flow IV as the amount of training data increases. The results can be seen in Figure~\ref{fig:convergence}.

\begin{align*}
    Z &= U_Z \\
    A &= 0.5 (Z + U_A) \\
    \Y &= \begin{pmatrix}
        A \\
        A
    \end{pmatrix} + \Ynoise \\
    \text{with}\qquad\qquad U_Z &\sim \Normal{0}{1} \\
    \begin{pmatrix}
        U_A \\
        U_{Y_1} \\
        U_{Y_2}
    \end{pmatrix}
    &\sim \Normal{\mathbf{0}}{\begin{pmatrix}
1 & \rho & \rho\\
\rho & 1 & \rho^2\\
\rho & \rho^2 & 1
\end{pmatrix}}, \quad \rho = 0.5 \\
\end{align*}

\subsubsection{Triangular Flow Matching}
\label{app:trianguar-flow-matching}

\begin{figure}[ht]
    \centering
    \includegraphics[width=0.5\linewidth]{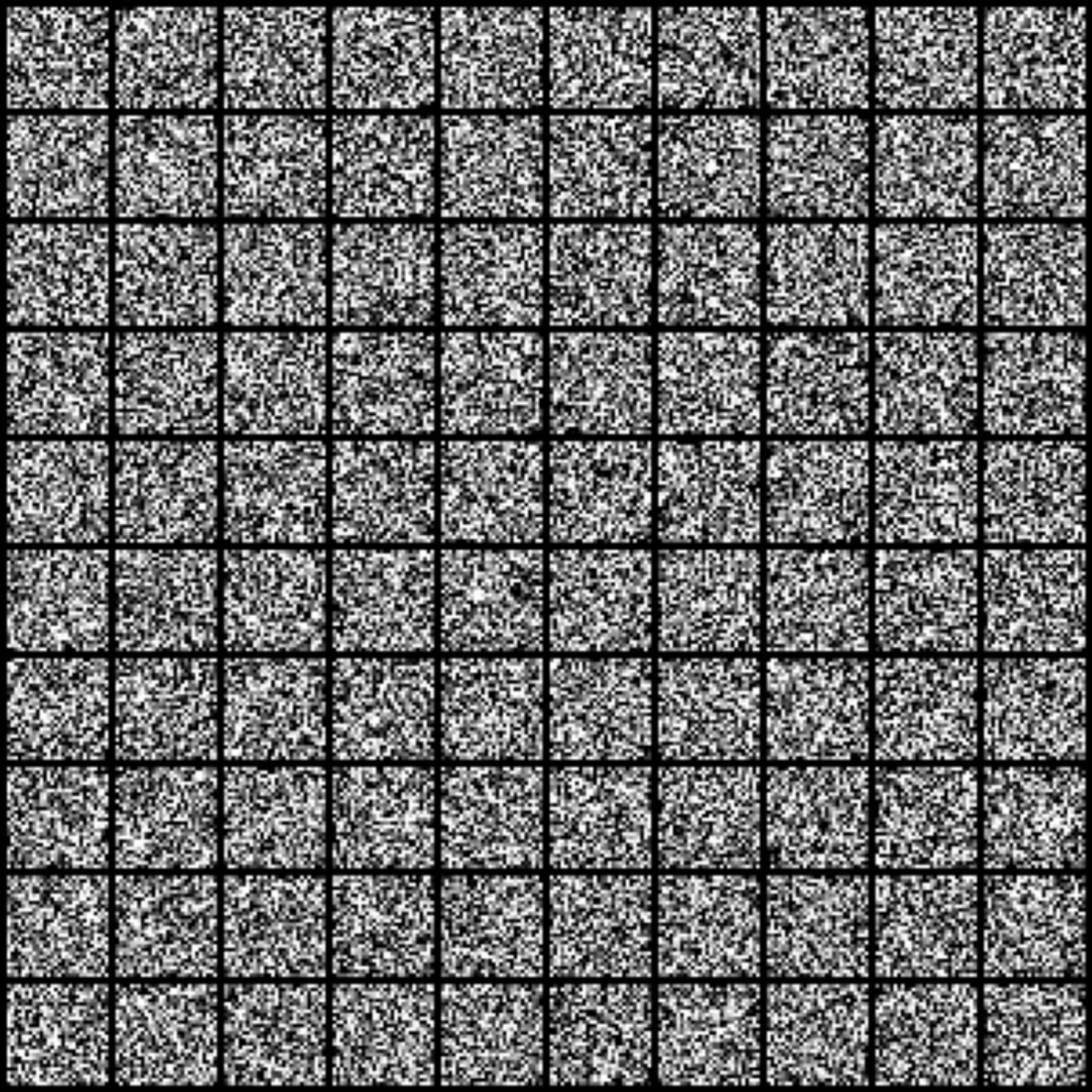}
    \caption{$10\times10$ grid of samples generated from a flow matching model with triangular \ac{MLP} as vector field network.}
    \label{fig:triangular-flow-matching}
\end{figure}

In Section~\ref{sec:high-dimensional-outcome} we present an experiment where we use flow matching with Flow IV but do not enforce that the outcome function has a triangular structure. We rerun the same experiment where we use a triangular \ac{MLP} network for the vector field. However, as the examples in Figure~\ref{fig:triangular-flow-matching} illustrates, the model failed to produce realistic looking images. Our assumption is that the \ac{MLP} is not flexible enough for the task. How the triangular structure can be enforced with more flexible network architectures remains an open question for future research.

\subsection{Training Details and Model Architecture}
\label{app:training-details-and-model-architecture}

Training the models in this study was performed on a RTX 3060 GPU with 12 GB of VRAM running on Linux. For the normalizing flows we used conditional quadratic rational spline transformations with one flow step and 32 spline bins for all experiments. The conditioning network had three hidden layers with 24 hidden units and we used a learning rate of 0.00544 and a batch size of 256 for the Dreher example. For the synthetic data examples, we used a network with two hidden layers with ten hidden units and a learning rate of 0.001. The flow matching model was a UNet with seven convolutional layers (batch size 128). The Deep IV model is a \ac{MLP} with two layers with ten hidden units and a learning rate of 0.001. For the GCFN we use a categorical control variable with 50 categories using two \acp{MLP} with two hidden layers with 100 hidden units for the encoder and decoder, followed by an outcome model using another 2-layer, 50-unit \ac{MLP}. The learning rates are all set to 0.01 for encoder, decoder, marginal, and outcome model. The training set for the synthetic data was of size 15,000 and the batch size used for all models was 500. We used early stopping on a validation set of size 9,000 for all models.
All details and the full replication code and models can be found in the repository \href{https://github.com/Marbr987/flow-iv}{https://github.com/Marbr987/flow-iv}.
%
%
%
%
\section{Justification of Assumption 1}
\label{app:justification-of-a1}

As stated in Section~\ref{sec:identifiability}, the triangularity assumption of Assumption~\ref{ass:triangular-monotonic} is justified for example when a proces unfolds over time. Such process is illustrated in the following figure.

\begin{figure}[ht]
\centering
\begin{tikzpicture}[
  node distance=1.5cm and 1.5cm,
  every node/.style={draw, circle, minimum size=1.4cm},
  every path/.style={->,>=stealth,thick},
  scale=2
]    
    \node[dashed] (eY1) at (-1.5, 1.2) {$U_{\y}^{(t_0)}$};
    \node[dashed] (eY2) at (0, 1.2) {$U_{\y}^{(t_1)}$};
    \node[dashed] (eY3) at (1.5, 1.2) {$U_{\y}^{(t_2)}$};
    \node (Y1) at (-1.5, 0) {$Y^{(t_0)}$};
    \node (Y2) at (0, 0) {$Y^{(t_1)}$};
    \node (Y3) at (1.5, 0) {$Y^{(t_2)}$};
    \node (A) at (-1.5, -1) {$\A$};
    
    \draw[] (A) -- (Y1);
    \draw[] (A) -- (Y2);
    \draw[] (A) -- (Y3);
    \draw[] (eY1) -- (Y1);
    \draw[] (eY1) -- (Y2);
    \draw[] (eY1) -- (Y3);
    \draw[] (eY2) -- (Y2);
    \draw[] (eY2) -- (Y3);
    \draw[] (eY3) -- (Y3);

\end{tikzpicture}
\caption{Causal graph illustrating how a triangular structure can arise from temporal processes.}
\label{fig:temporal-dag}
\end{figure}

%
%
%
%
\section{Algorithms}
\label{app:algorithms}

Next we present the training procedures for Flow IV with normalizing flows and for Flow IV with flow matching as pseudocode.


\begin{algorithm}[H]
\DontPrintSemicolon

\hrulefill
\caption{Flow IV Training with Normalizing Flows}\label{alg:flowiv-nf-training}
\vspace{-1.5ex}
\hrulefill

\KwIn{Training dataset $\mathcal{D}=\{(\z_i,\a_i,\y_i)\}_{i=1}^{\ell}$\\ \textbf{Hyperparameters:} Learning rate $\eta$, number of iterations $T$}
\KwOut{Trained parameters $\theta=(\theta_{\z},\theta_{\a},\theta_{\y},\theta_{\noisehat})$}

Randomly initialize parameters $\theta$\;

\For{$t \leftarrow 1$ \KwTo $T$}{
  Sample a minibatch $\mathcal{B}\subset \mathcal{D}$\;
  
  $\log \mathcal{L} \leftarrow 0$\;

  \ForEach{$(\z_i,\a_i,\y_i)\in \mathcal{B}$}{
    $(\zhatnoise,\ahatnoise,\yhatnoise) \leftarrow g^{-1}(\z_i,\a_i,\y_i;\theta)$\;
    
    $(\mathbf{\varepsilon}_{\a},\mathbf{\varepsilon}_{\y}) \leftarrow h^{-1}(\ahatnoise,\yhatnoise;\theta_{\tilde{u}})$\;

    $\log \mathcal{L} \leftarrow \log \mathcal{L}
      + \log \varphi(\zhatnoise)
      + \log \varphi(\mathbf{\varepsilon}_{\a})
      + \log \varphi(\mathbf{\varepsilon}_{\y})
      + \log\left|\det J_{h^{-1}}(\ahatnoise,\yhatnoise;\theta_{\tilde{u}})\right|
      + \log\left|\det J_{g^{-1}}(z_i,a_i,y_i;\theta)\right|$\;
  }

  $\theta \leftarrow \theta + \eta \nabla_{\theta}\log \mathcal{L}$ \tcp*[r]{gradient ascent (MLE)}
}

\vspace{-1.5ex}
\hrulefill
\end{algorithm}

\begin{algorithm}[H]
\DontPrintSemicolon

\hrulefill
\caption{Flow IV Training with Flow Matching}\label{alg:flowiv-cfm-training}
\vspace{-1.5ex}
\hrulefill

\KwIn{Training dataset $\mathcal{D}=\{(\z_i,\a_i,\y_i)\}_{i=1}^{\ell}$, \\ \textbf{Hyperparameters:} Learning rates $\eta_1, \eta_2, \eta_3, \eta_4$, number of iterations $T_1$, $T_2$, noise schedule $\sigma(t, \y)$, conditional mean function $\mu(t, \y)$}
\KwOut{Trained parameters $\theta=(\theta_{\z},\theta_{\a},\theta_{\y},\theta_{\noisehat})$}

Randomly initialize parameters $\theta$\;

\tcp*[l]{Step 1: Optimize normalizing flows $g_{\z}$ and $g_{\a}$}
\For{$t \leftarrow 1$ \KwTo $T_1$}{
  Sample a minibatch $\mathcal{B}\subset \mathcal{D}$\;
  
  $\log \mathcal{L} \leftarrow 0$\;

  \ForEach{$(\z_i,\a_i,\y_i)\in \mathcal{B}$}{
    $\zhatnoise \leftarrow g_{\z}^{-1}(\z_i;\theta_{\z})$\;
    
    $\ahatnoise \leftarrow g_{\a}^{-1}(\z_i,\a_i;\theta_{\a})$\;

    $\log \mathcal{L} \leftarrow \log \mathcal{L}
      + \log \varphi(\zhatnoise)
      + \log \varphi(\ahatnoise)
      + \log\left|\det J_{(g_{\z}^{-1}, g_{\a}^{-1})}(z_i,a_i;\theta)\right|$\;
  }

  $\theta_{\z} \leftarrow \theta_{\z} + \eta_1 \nabla_{\theta_{\z}}\log \mathcal{L}$ \; \tcp*[r]{gradient ascent (MLE)}

  $\theta_{\a} \leftarrow \theta_{\a} + \eta_2 \nabla_{\theta_{\a}}\log \mathcal{L}$
}

\tcp*[l]{Step 2: Optimize normalizing flow $h$ and vector field $v$}
\For{$t \leftarrow 1$ \KwTo $T_2$}{
  Sample a minibatch $\mathcal{B}\subset \mathcal{D}$\;
  
  $\mathcal{L} \leftarrow 0$\;

  \ForEach{$(\z_i,\a_i,\y_i)\in \mathcal{B}$}{    
    $\ahatnoise \leftarrow g_{\a}^{-1}(\z_i,\a_i;\theta_{\a})$\;

    Sample $\mathbf{\varepsilon}_{\y} \sim \Normal{0}{\mathbf{I}}$\;

    $\yhatnoise \leftarrow h(\ahatnoise, \mathbf{\varepsilon}_{\y}; \theta_{\noisehat})$\;

    Sample $t \sim \Unif [0, 1]$\;

    $\y_t \leftarrow \mu(t, \y_i) + \sigma(t, \y_i) \yhatnoise$

    $v_{\text{cond}} \leftarrow \frac{\sigma'(t, \y_i)}{\sigma(t, \y_i)} (\y_t - \mu(t, \y_i)) + \mu'(t, \y_i)$\;

    $\mathcal{L} \leftarrow \mathcal{L} + \left( v_{\text{cond}} - v(t, a, y_t; \theta_{\y}) \right)^2$\;
  }

  $\theta_{\noisehat} \leftarrow \theta_{\noisehat} - \eta_3 \nabla_{\theta_{\noisehat}}\mathcal{L}$ \; \tcp*[r]{gradient descent}

  $\theta_{\y} \leftarrow \theta_{\y} - \eta_4 \nabla_{\theta_{\y}} \mathcal{L}$
}

\vspace{-1.5ex}
\hrulefill
\end{algorithm}

\end{document}